\documentclass[journal,compsoc]{IEEEtran}
\usepackage[T1]{fontenc}
\usepackage[latin9]{inputenc}
\usepackage{color}
\usepackage{array}
\usepackage{rotating}
\usepackage{multirow}
\usepackage{amsmath}
\usepackage{amssymb}
\usepackage{graphicx}
\usepackage{setspace}
\usepackage{algorithm}
\usepackage[unicode=true,
 bookmarks=false,
 breaklinks=true,pdfborder={0 0 1},backref=section,colorlinks=false]
 {hyperref}

\makeatletter

\providecommand{\tabularnewline}{\\}

  \usepackage[nocompress]{cite}

\usepackage{times}
\usepackage{epsfig}

\usepackage{array}
\usepackage{textcomp}
\usepackage{multirow}
\usepackage{xcolor}

\usepackage{amsthm}
\theoremstyle{definition}
\newtheorem{definition}{Definition}[section]

\DeclareTextFontCommand{\emph}{\em}
 
\newcommand{\ie}{\textit{i}.\textit{e}.}
\newcommand{\eg}{\textit{e}.\textit{g}.}
\newcommand{\etal}{\textit{et al}.}

\newcommand{\leon}[1]{\textcolor{black}{#1}}

\makeatother

\begin{document}
\title{Vocabulary-informed Zero-shot and Open-set Learning }  

\author{ Yanwei~Fu,   Xiaomei Wang, Hanze Dong, Yu-Gang Jiang,   Meng Wang, Xiangyang Xue, Leonid Sigal  
\IEEEcompsocitemizethanks{ 
\IEEEcompsocthanksitem Yanwei~Fu and Hanze Dong are with the School of Data Science, Fudan University, Shanghai. Email: \{yanweifu,hzdong15\}@fudan.edu.cn. 
 \IEEEcompsocthanksitem Xiaomei Wang, Yu-Gang Jiang  and Xiangyang Xue are with the School of Computer Science, Shanghai Key Lab of Intelligent Information Processing, Fudan University. Email: \{17110240025,ygj,xyxue\}@fudan.edu.cn. 
 \IEEEcompsocthanksitem  Yu-Gang Jiang is the  corresponding author. Yanwei~Fu is also with AITRICS.
\IEEEcompsocthanksitem Meng Wang is with the School of Computer and Information Science, Hefei University of Technology, Hefei, China. Email: eric.wangmeng@gmail.com. \IEEEcompsocthanksitem  Leonid Sigal is with the Department of Computer Science, University of British Columbia, BC, Canada. Email: lsigal@cs.ubc.ca.  }\thanks{}  } 

\IEEEcompsoctitleabstractindextext{  

\begin{abstract}  
Despite significant progress in object categorization, in recent years, a number of important challenges remain; mainly, the ability to learn from limited labeled data and to recognize object classes within large, potentially open, set of labels. Zero-shot learning is one way of addressing these challenges, but it has only been shown to work with limited sized class vocabularies and typically requires separation between supervised and unsupervised classes, allowing former to inform the latter but not vice versa. We propose the notion of vocabulary-informed learning to alleviate the above mentioned challenges and address problems of supervised, zero-shot, generalized zero-shot and open set recognition using a unified framework. Specifically, we propose a weighted maximum margin framework for semantic manifold-based recognition that incorporates distance constraints from (both supervised and unsupervised) vocabulary atoms. \leon{Distance constraints ensure} that labeled samples are projected \leon{closer} to their correct prototypes, in the embedding space, than to others.  We \leon{illustrate} that resulting model shows improvements in supervised, zero-shot, generalized zero-shot, and large open set recognition, with up to 310K class vocabulary on Animal with Attributes and ImageNet datasets.   
\end{abstract}  

\begin{IEEEkeywords}
Vocabulary-informed learning, Generalized zero-shot learning, Open-set recognition, Zero-shot learning.  
\end{IEEEkeywords} } 
\maketitle

\section{Introduction\label{sec:Introduction}}

Object recognition, more specifically object categorization, has
seen unprecedented advances in recent years with development of convolutional
neural networks (CNNs) \cite{krizhevsky2012imagenet}. However, most successful
recognition models, to date, are formulated as supervised learning
problems, in many cases requiring hundreds, if not thousands, labeled
instances to learn a given concept class \cite{deng2009imagenet}.
This exuberant need for large labeled instances has limited recognition
models to domains with hundreds to thousands of classes. Humans, on
the other hand, are able to distinguish beyond $30,000$ basic level
categories \cite{biederman1987recognition}. Even more impressive is the
fact that humans can learn from few examples, by effectively leveraging
information from other object category classes, and even recognize
objects without ever seeing them (\eg, by reading about them on the
Internet). This ability has spawned the research in few-shot and zero-shot
learning.

Zero-shot learning (ZSL) has now been widely studied in a variety
of research areas including neural decoding of fMRI images \cite{palatucci2009zero},
character recognition~\cite{larochelle2008zero}, face verification
\cite{kumar2009attribute}, object recognition \cite{lampert2014attribute}, and
video understanding~\cite{fu2013learning,wu2016harnessing}.
Typically, zero-shot learning approaches aim to recognize instances
from the unseen or unknown testing {\em target} categories by transferring
information through intermediate-level semantic representations,
from known observed {\em source} (or auxiliary) categories for
which many labeled instances exist. In other words, supervised classes/instances,
are used as context for recognition of classes that contain no visual
instances at training time, but that can be put in some correspondence
with supervised classes/instances. Therefore, a general experimental
setting of ZSL is that the classes in target and source (auxiliary)
dataset are disjoint. Typically, the learning is done on the source
dataset and then information is transferred to the target dataset,
with performance measured on the latter.

This setting has a few important drawbacks: (1) it assumes that target
classes cannot be mis-classified as source classes and vice versa;
this greatly and unrealistically simplifies the problem; (2) the target
label set is often relatively small, between ten \cite{lampert2014attribute}
and several thousand unknown labels \cite{frome2013devise}, compared
to at least $30,000$ entry level categories that humans can distinguish;
(3) large amounts of data in the source (auxiliary) classes are required,
which is problematic as it has been shown that most object classes
have very few instances (long-tailed distribution of objects in the
world \cite{torralba200880}); and (4) the vast open set vocabulary \leon{and
corresponding} semantic knowledge, defined as part of ZSL~\cite{palatucci2009zero},
is not leveraged in any way to inform the learning or source class
recognition.

\begin{figure*}
\begin{centering}
\includegraphics[scale=0.36]{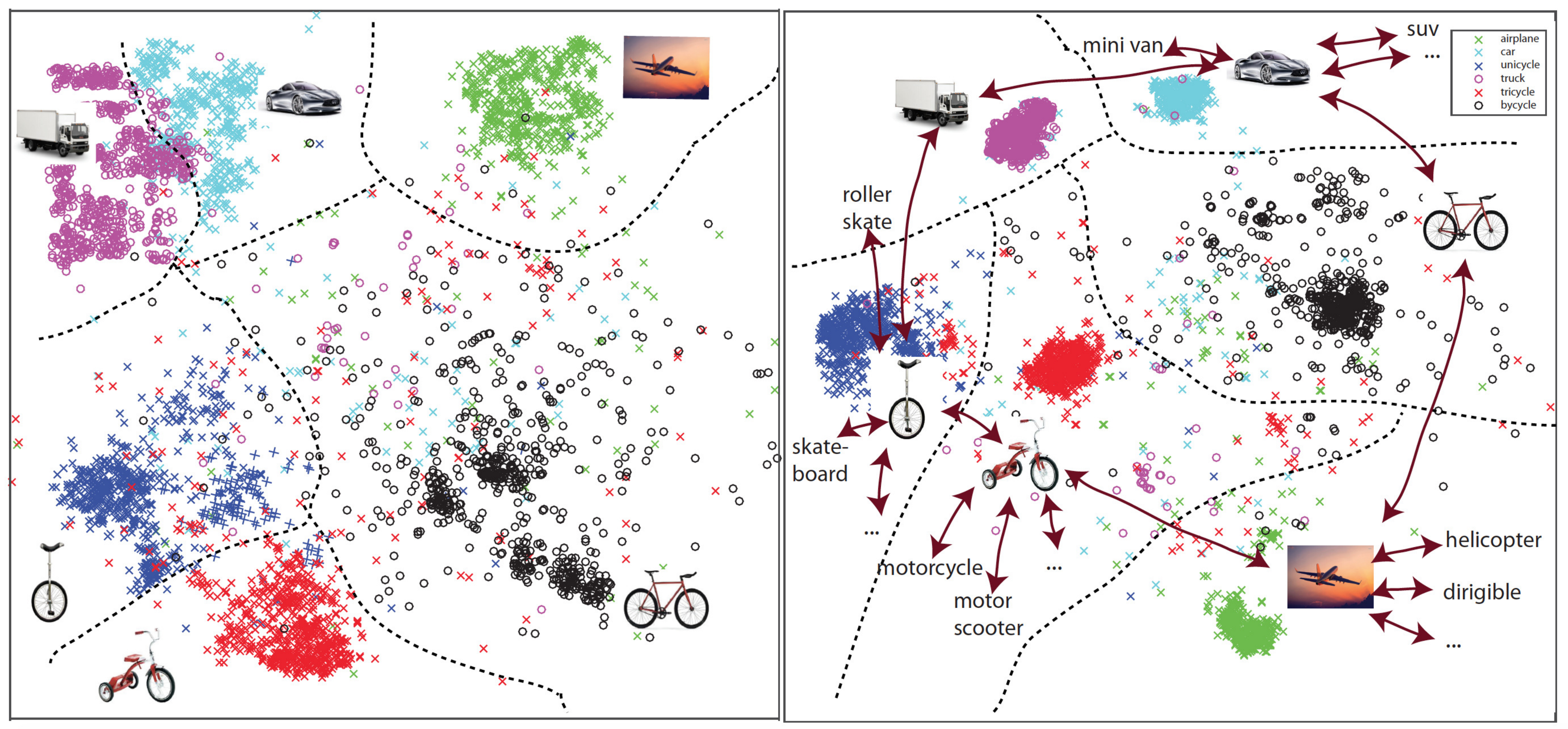} 
\par\end{centering}
\caption{\label{fig:intro} \textbf{Illustration of the semantic embeddings}
learned (left) using support vector regression (\textbf{SVR}) and
(right) using the proposed vocabulary-informed learning (\textbf{Deep WMM-Voc})
approach. In both cases, t-SNE visualization is used to illustrate
samples from $4$ source/auxiliary classes (denoted by $\times$)
and $2$ target/zero-shot classed (denoted by $\circ$) from the ImageNet
dataset. Decision boundaries, illustrated by dashed lines, are drawn
by hand for visualization. 
\textcolor{black}{
The large margin constraints, both among the source/target classes and the
external vocabulary atoms, are denoted by arrows and words on the right. 
Note that the WMM-Voc approach on the right 
leads to a better embedding with more compact and separated
classes (\eg, see {\em truck} and {\em car} or {\em unicycle}
and {\em tricycle}). }}
\end{figure*}

A few works recently looked at resolving (1) through class-incremental
learning \cite{Scheirer_2013_TPAMI,socher2013zero} or generalized
zero-shot learning (G-ZSL) \cite{chao2016empirical,palatucci2009zero} which
are designed to distinguish between seen (source) and unseen (target)
classes at the testing time and apply an appropriate model -- supervised
for the former and ZSL for the latter. However, (2)--(4) remain largely
unresolved. In particular, while (2) and (3) are artifacts of the
ZSL setting, (4) is more fundamental; \leon{\eg, a recent study \cite{huth2012continuous}
argues that concepts, in our own brains, are represented in the form
of a continuous semantic space mapped smoothly across the cortical
surface.} For example, consider learning about a {\em car} by
looking at image instances in \leon{Figure}~\ref{fig:intro}. Not
knowing that other motor vehicles exist in the world, one may be tempted
to call anything that has 4-wheels a {\em car}. As a result, the
zero-shot class {\em truck} may have a large overlap with the {\em
car} class (see \leon{Figure} \ref{fig:intro} (left)). However,
imagine knowing that there also exist many other motor vehicles (trucks,
mini-vans, {\em etc}). Even without having visually seen such objects,
the very basic knowledge that they {\em exist} in the world and
are closely related to a {\em car} should, in principle, alter
the criterion for recognizing instance as a {\em car} (making the
recognition criterion stricter in this case). Encoding this in our
vocabulary-informed learning model results in better separation among
classes (see~\leon{Figure}~\ref{fig:intro}~(right)).

To tackle the limitations of ZSL and towards the goal of generic recognition,
we propose the idea of\textit{ vocabulary-informed learning}. Specifically,
assuming we have few labeled training instances and a large, potentially
open set, vocabulary/semantic dictionary (along with textual sources
from which statistical semantic relations among vocabulary atoms can
be learned), the task of vocabulary-informed learning is to learn
a \leon{unified} model that utilizes this semantic dictionary to
help train better classifiers for observed (source) classes and unobserved
(target) classes in supervised, zero-shot, generalized zero-shot,
and open set image recognition settings.

\leon{ In particular, we formulate Weighted Maximum Margin Vocabulary-informed  Embedding
(WMM-Voc), which learns a joint embedding 
for visual features and semantic words. In this formulation,
two maximum margin sets of constraints are simultaneously optimized.
The first set ensures that labeled training visual instances, belonging
to a particular class, project close to semantic word vector prototype
corresponding to the class name in the embedding space. The second
set ensures that these instances are closer to the correct class word
vector prototype than to any of the incorrect ones in the embedding
space; including those that may not contain training data (\ie, zero-shot).
The constraints in the first set further take into the account the
distribution of training samples for each class, and nearby classes,
to dynamically set appropriate margins. 
In other words, for some classes the distance, between the projected
training sample and the word vector prototype, is explicitly penalized
more (or less) than for others. This weighting is derived using extreme
values theory.}


\noindent \textbf{Contributions:} Our main contribution is to propose
a novel paradigm for potentially open set image recognition:\textit{
vocabulary-informed learning} (\textbf{Voc}), which is capable of
utilizing vocabulary over unsupervised items, during training, to
improve recognition. \leon{We extend the model initially proposed
by us in a conference paper \cite{fu2016semi} to include class-specific
weighting in the data term, as well as the ability to run the models
as an end-to-end network.} Particularly, classification is done through the
nearest-neighbor distance to class prototypes in the semantic embedding
space. Semantic embedding is learned subject to 
constraints ensuring that labeled images
project into semantic space such that they end up closer to the correct
class prototypes than to incorrect ones (whether those prototypes
are part of the source or target classes). We show that word embedding
(word2vec) can be used effectively to initialize the semantic space.
Experimentally, we illustrate that through this paradigm: we can achieve very competitive supervised (on source classes), ZSL (on target classes)
and G-ZSL performance, as well as open set image recognition performance
with a large number of unobserved vocabulary entities (up to $300,000$);
effective learning with few samples is also illustrated. Critically,
our models can be directly utilized in G-ZSL scenario and still has
much better results than the baselines.

\section{Related Work}

Our \leon{model} belongs to \leon{a class of} transfer
learning \leon{approaches} \cite{pan2009transfer_survey}, also
\leon{sometimes} called meta-learning \cite{vilalta2002perspective} or
learning to learn \cite{thrun1996learning}. The key idea of transfer
learning is to transfer the knowledge from previously learned categories
to recognize new categories with no \leon{training} example\leon{s}
(zero-shot learning \cite{lampert2014attribute,rohrbach2011evaluating}), few
examples (one-shot learning \cite{tommasi2009more,feifei2003unsup_1s_objcat_learn})
or from vast open set vocabulary \cite{fu2016semi}. The process of knowledge
transfer can be done by sharing features \cite{bart2005cross,hertz2006learning,fleuret2006pattern,amit2007uncovering,wolf2005robust,torralba2007sharing},
semantic attributes \cite{lampert2014attribute,rohrbach2013transfer,rohrbach2010helps},
or contextual information \cite{torralba2010using}.

Visual-semantic embedding\leon{s} \leon{have} been widely used
for transfer learning. \leon{Such models} 
embed visual feature\leon{s} in\leon{to a} semantic space by learning
projections \leon{of} different forms. \leon{Examples include}
WSABIE \cite{weston2011wsabie}, ALE \cite{akata2015label},
SJE \cite{akata2015evaluation}, DeViSE \cite{frome2013devise}, SVR \cite{farhadi2009describing,lampert2014attribute},
kernel embedding 
\cite{hertz2006learning} \leon{and} Siamese networks \cite{koch2015siamese}.

\subsection{Open-set Recognition }

\noindent The term ``open set recognition'' was initially defined
in \cite{scheirer2014probability,Scheirer_2013_TPAMI} and formalized
in \cite{bendale2015towards,sattar2015prediction,bendale-boult-cvpr2016}
which mainly aim at identifying whether an image belongs to a seen
or unseen classes. \leon{The problem} is also known as class-incremental
learning. However, none of \leon{these methods} can further identify
classes for unseen instances. \textcolor{black}{The exceptions are
\cite{norouzi2013zero,frome2013devise} which augment zero-shot
(unseen) class labels with source (seen) labels in some of their experimental
settings.} Similarly, we define the \emph{open set image recognition}
as the problems of recognizing the class name of an image from a potentially
very large open set vocabulary (including, but not limited to source
and target labels). Note that methods like \cite{scheirer2014probability,Scheirer_2013_TPAMI}
are orthogonal but potentially useful here -- it is still worth identifying
seen or unseen instances to be recognized with different label sets.
Conceptually similar, but different in formulation and task, open-vocabulary
object retrieval~\cite{guadarrama2014open} focused on retrieving objects
using natural language open-vocabulary queries.

\subsection{One-shot Learning}

\noindent While most of machine learning-based object recognition
algorithms require a large amount of training data, one-shot learning
\cite{Fei-Fei:2006:OLO:1115692.1115783} aims to learn object classifiers from one,
or very few examples. To compensate for the lack of training instances
and enable one-shot learning, {\em knowledge} must be transferred
from other sources, for example, by sharing features \cite{bart2005cross},
semantic attributes \cite{fu2013learning,lampert2014attribute,rohrbach2013transfer,rohrbach2010helps},
or contextual information \cite{torralba2010using}. However,
none of the previous work had used the open set vocabulary to help learn
the object classifiers.

\subsection{Zero-shot Learning}

Zero-shot Learning (ZSL) aims to recognize novel classes with no training instance by transferring
{\em knowledge} from source classes. ZSL was first explored with
use of attribute-based semantic representations~\cite{farhadi2009describing,fu2012attribute,fu2013learning,fu2015transductive,kumar2009attribute,parikh2011relative}.
This required pre-defined attribute vector prototypes for each class,
which is costly to obtain for a large-scale dataset. Recently, semantic word
vectors were proposed as a way to embed any class name without human
annotation effort; they can therefore serve as an alternative semantic
representation \cite{akata2015evaluation,frome2013devise,fu2015zero,norouzi2013zero}
for ZSL. Semantic word vectors are learned from large-scale text corpus
by language models, such as word2vec \cite{mikolov2013distributed}
or GloVec \cite{pennington2014glove}. However, most of the previous work only use
word vectors as semantic representations in ZSL setting, but have
neither (1) utilized semantic word vectors explicitly for learning
better classifiers; nor (2) for extending ZSL setting towards open
set image recognition. A notable exception is \cite{norouzi2013zero}
which aims to recognize 21K zero-shot classes given a modest vocabulary
of 1K source classes; we explore vocabularies that are up to an order
of the magnitude larger -- 310K.

Generalized zero-shot recognition (G-ZSL) \cite{chao2016empirical}
relaxed the problem setup of conventional zero-shot learning by considering
the training classes in the recognition step. Chao \emph{et al.} \cite{chao2016empirical}
investigated the G-ZSL task and found that it is less effective to
directly extend the existing zero-shot learning algorithms \leon{to
deal with G-ZSL setting}. Recently, Xian \emph{et al.} \cite{palatucci2009zero}
systematically compared the evaluation settings for ZSL and G-ZSL.
Comparing against existing ZSL models, which are inferior in the G-ZSL
scenario, we show that our vocabulary-informed frameworks can be directly
utilized for G-ZSL and achieve very competitive performance.

\subsection{Visual-semantic Embedding}

\noindent Mapping between visual features and semantic entities has
been explored in three ways: (1) directly learning the embedding by
regressing from visual features to the semantic space using Support
Vector Regressors (SVR) \cite{farhadi2009describing,lampert2014attribute}
or neural network \cite{socher2013zero}; (2) projecting visual features
and semantic entities into a common {\em new} space, such as SJE
\cite{akata2015evaluation}, WSABIE \cite{weston2011wsabie},
ALE \cite{akata2015label}, DeViSE \cite{frome2013devise}, and
CCA \cite{fu2014transductive,fu2015transductive}; (3) learning
the embeddings by regressing from the semantic space to visual features,
including \cite{kodirov2017semantic,embedding_cvpr17,long2017zero,changpinyo2017predicting}.

In contrast to other embedding methods, our model trains a better visual-semantic embedding
from only few training instances with the help of a large amount of
open set vocabulary items (using a maximum margin strategy). Our formulation
is inspired by the unified semantic embedding model of \cite{hwang2014unified},
however, unlike \cite{hwang2014unified}, our formulation is built on
word vector representation, contains a data term, and incorporates
constraints to unlabeled vocabulary prototypes.

\section{Vocabulary-informed Learning}

\subsection{Problem setup }

\noindent Assume a labeled source dataset $\mathcal{D}_{s}=\{\mathbf{x}_{i},z_{i}\}_{i=1}^{N_{s}}$
of $N_{s}$ samples, where $\mathbf{x}_{i}\in\mathbb{R}^{p}$ is the
image feature representation of image $i$ and $z_{i}\in\mathcal{W}_{s}$
is a class label taken from a set of English words or phrases $\mathcal{W}$;
consequently, $|\mathcal{W}_{s}|$ is the number of source classes.
Further, suppose another set of class labels for target classes $\mathcal{W}_{t}$,
\leon{also taken from $\mathcal{W}$}, such that $\mathcal{W}_{s}\cap\mathcal{W}_{t}=\emptyset$,
for which no labeled samples are available. We note that potentially
$|\mathcal{W}_{t}|>>|\mathcal{W}_{s}|$.

Given a new test image feature vector $\mathbf{x}^{*}$ the goal is
then to learn a function $z^{*}=f(\mathbf{x}^{*})$, using all available
information, that predicts a class label $z^{*}$. Note that the form
of the problem changes drastically depending on \leon{the} label
set assumed for $z^{*}$: 
\begin{itemize}
\item Supervised learning: $z^{*}\in\mathcal{W}_{s}$; 
\item Zero-shot learning: $z^{*}\in\mathcal{W}_{t}$ ; 
\item Generalized zero-shot learning: $z^{*}\in\{\mathcal{W}_{s},\mathcal{W}_{t}\}$; 
\item Open set recognition: $z^{*}\in\mathcal{W}$. 
\end{itemize}
\leon{Note that open set recognition is similar to generalized zero-shot
learning, however, in open set setting additional {\em distractor}
classes that do not exist in either source or target datasets are present.} We
posit that a single unified $f(\mathbf{x}^{*})$ can be learned for
all 
cases. We formalize the definition of vocabulary-informed learning
(Voc) as follows:

\theoremstyle{definition} \begin{definition}{ \em Vocabulary-informed
Learning (Voc): } is a learning setting that makes use of complete
vocabulary data ($\mathcal{W}$) during training. Unlike a more traditional
ZSL that typically makes use of the vocabulary (\eg, semantic embedding)
at test time, Voc utilizes exactly the same data during training.
Notably, Voc requires no additional annotations or semantic knowledge;
it simply shifts the burden from testing to training, leveraging the
vocabulary to learn a better model. \end{definition}

The vocabulary $\mathcal{W}$ can be represented by semantic embedding space
learned by word2vec~\cite{mikolov2013distributed} or GloVec~\cite{pennington2014glove}
on large-scale corpus; each vocabulary entity $w\in\mathcal{W}$ is
represented as a distributed semantic vector $\mathbf{u}\in\mathbb{R}^{d}$.
Semantics of embedding space help with knowledge transfer among classes,
and allow ZSL, G-ZSL and open set image recognition. Note that such
semantic embedding spaces are equivalent to the ``semantic knowledge
base'' for ZSL defined in \cite{palatucci2009zero} and hence
make it appropriate \leon{to} use \leon{Vocabulary-informed Learning}
in ZSL. 

\subsection{Learning Embedding and Recognition}

Assuming we can learn a mapping $g:\mathbb{R}^{p}\rightarrow\mathbb{R}^{d}$,
from image features to this semantic space, recognition can be carried
out using simple nearest neighbor distance, \eg, $f(\mathbf{x}^{*})=car$
if $g(\mathbf{x}^{*})$ is closer to $\mathbf{u}_{car}$ than to any
other word vector; $\mathbf{u}_{j}$ in this context can be interpreted
as the prototype of the class $j$. \textcolor{black}{Essentially,
the attribute or semantic word vector of the class name can be taken
as the class prototype \cite{fu2018recent}}. The core question
is then how to learn the mapping $g(\mathbf{x})$ and what form of
inference is optimal in the semantic space. For learning we propose
the discriminative maximum margin criterion that ensures that labeled
samples $\mathbf{x}_{i}$ project closer to their corresponding class
prototypes $\mathbf{u}_{z_{i}}$ than to any other prototype $\mathbf{u}_{i}$
in the open set vocabulary $i\in\mathcal{W}\setminus z_{i}$. 


\noindent \textbf{\noindent Learning Embedding:} To learn the function
$f(\mathbf{x})$, one needs to establish the correspondence between
visual feature space and semantic space. Particularly, in the training
step, each image sample $\mathbf{x}_{i}$ is regressed towards its corresponding
class prototype $\mathbf{u}_{z_{i}}$ by minimizing 
\begin{equation}
W=\arg\min_{W}\sum\limits _{i=1}^{Ns}L\left(\mathbf{x}_{i},\mathbf{u}_{z_{i}}\right)+\lambda\parallel W\parallel_{F}^{2}\label{eq:initial_regression}
\end{equation}
where $L\left(\mathbf{x}_{i},\mathbf{u}_{z_{i}}\right)=\Vert g\left(\mathbf{x}_{i}\right)-\mathbf{u}_{z_{i}}\Vert_{2}^{2}$
; and $g:\mathbb{R}^{p}\rightarrow\mathbb{R}^{d}$ is the mapping
from image features to semantic space; $\parallel\cdot\parallel_{F}$
indicates the Frobenius Norm. If $g\left(\mathbf{x}\right)=W^{T}\mathbf{x}$
is a linear mapping, we have the closed form solution for Eq.~(\ref{eq:initial_regression}).
The loss function in Eq.~(\ref{eq:initial_regression}) can be interperted
as a variant of SVR embedding. \leon{However, this is too limiting.}
To learn the linear embedding matrix $W$, we introduce and discuss two
sets of methods in Section~\ref{subsec:Maximum-Margin-Voc} and 
Section~\ref{subsec:Extreme-Value-Voc}.


\noindent \textbf{\noindent Recognition:} The recognition step can
\leon{be formulated using the} 
nearest neighbor classifier. Given a testing instance \leon{$\mathbf{x}^{\star}$},

\begin{equation}
z^{\star}=\arg\min_{i}\left\Vert W^{T}\mathbf{x}^{\star}-\mathbf{u}_{i}\right\Vert _{2}^{2}.\label{eq:recognition}
\end{equation}

\noindent \leon{Eq.~(\ref{eq:recognition})} 
measures the distance between predicted vector \leon{and}
the class prototypes in the semantic space. In terms of different label set,
we can do supervise, zero-shot, generalized zero-shot or open set
recognition without modifications.

In particular, we explore a simple variant of Eq.~(\ref{eq:recognition})
to classify the testing instance $\mathbf{x}^{\star}$, 

\begin{equation}
z^{*}=\arg\min_{i}\parallel W^T\mathbf{x}^{*}-\omega\left(\mathbf{u}_{i}\right)\parallel_{2}^{2},\label{ZSL_classifier}
\end{equation}

\noindent
where the Nearest Neighbor (NN) classifier measures distance
between the predicted semantic vectors and a function of prototypes in the semantic
space, \eg, $\omega\left(\mathbf{u}_{i}\right)=\mathbf{u}_{i}$ is equivalent to Eq~(\ref{eq:recognition}).
In practice, we employ semantic vector prototype averaging to define $\omega\left(\cdot\right)$. 
 For example, sometimes, there might be more than one positive prototype, such as \emph{pig}, \emph{pigs} and \emph{hog}. 
 In such the circumstance, choosing the most likely prototype and using NN may not be sensible, hance we 
 introduce the averaging strategy to consider more prototypes for robustness. Note that this strategy
is known as Rocchio algorithm in information retrieval. 
Rocchio algorithm
is a method for relevance feedback that uses more relevant instances
to update the query for better recall and possibly precision
in the vector space (Chap 14 in \cite{manning2010introduction}).
It was first suggested for use in ZSL in \cite{fu2013learning};
more sophisticated algorithms \cite{fu2014transductive,rohrbach2013transfer}
are also possible.

\subsection{Maximum Margin Voc Embedding (MM-Voc)\label{subsec:Maximum-Margin-Voc}}

\noindent The maximum margin vocabulary-informed embedding learns
the mapping $g(\mathbf{x}):\mathbb{R}^{p}\rightarrow\mathbb{R}^{d}$,
from low-level features $\mathbf{x}$ to the semantic word space by
utilizing maximum margin strategy. Specifically, consider $g(\mathbf{x})=W^{T}\mathbf{x}$,
where\footnote{Generalizing to a kernel version is straightforward, see \cite{additiveKernel}.}
$W\subseteq\mathbb{R}^{p\times d}$. Ideally we want to estimate $W$
such that $\mathbf{u}_{z_{i}}=W^{T}\mathbf{x}_{i}$ for all labeled
instances in $\mathcal{D}_{s}$. Note that we would obviously want
this to hold for instances belonging to unobserved classes as well,
but we cannot enforce this explicitly in the optimization as we have
no labeled samples for them.

\noindent 
\textbf{Data Term:} The easiest way to enforce the above
objective is to minimize Euclidian distance between sample projections
and appropriate prototypes in the embedding space, 
\begin{equation}
D\left(\mathbf{x}_{i},\mathbf{u}_{z_{i}}\right)=\left\Vert W^{T}\mathbf{x}_{i}-\mathbf{u}_{z_{i}}\right\Vert _{2}^{2}.\label{eq:data-embedding}
\end{equation}

\noindent Where we need to minimize this term with respect to each
instance $\left(\mathbf{x}_{i},\mathbf{u}_{z_{i}}\right)$, where
$z_{i}$ is the class label of $\mathbf{x}_{i}$ in $\mathcal{D}_{s}$.
Such embedding is also known, in the literature, as data
embedding \cite{hwang2014unified} or compatibility function \cite{akata2015evaluation}.

To make the embedding more comparable to support vector regression
(SVR), we employ the maximal margin strategy -- $\epsilon-$insensitive
smooth SVR ($\epsilon-$SSVR) \cite{lee2005epsilon} in Eq. (\ref{eq:initial_regression}).
That is,

\begin{equation}
\mathcal{L}\left(\mathbf{x}_{i},\mathbf{u}_{z_{i}}\right)=\mathcal{L}_{\epsilon}\left(\mathbf{x}_{i},\mathbf{u}_{z_{i}}\right)+\lambda\parallel W\parallel_{F}^{2}\label{eq:SSVR}
\end{equation}

\noindent where $\mathcal{L}_{\epsilon}\left(\mathbf{x}_{i},\mathbf{u}_{z_{i}}\right)=\mathbf{1}^{T}\mid\xi\mid_{\epsilon}^{2}$;
$\lambda$ is regularization coefficient. 
\begin{equation}
\left(\left|\xi\right|_{\epsilon}\right)_{j}=\mathrm{max}\left\{ 0,\left|W_{\star j}^{T}\mathbf{x}_{i}-\left(\mathbf{\mathbf{u}}_{z_{i}}\right)_{j}\right|-w_{z_{i}}\cdot\epsilon\right\} \label{eq:epsilon}
\end{equation}
$|\xi|_{\epsilon}\in\mathbb{R}^{d}$; $\left(\right)_{j}$ indicates
the $j$-th value of corresponding vector; $W_{\star j}$ is the $j$-th
column of $W$, and $w_{z_{i}}$ is the scaling weight \leon{derived
from the density of class $z_{i}$ and it's neighboring classes.}
In our conference version \cite{fu2016semi}, equal weight $w_{z_{i}}$
is used for \leon{all} class\leon{es}. Here we notice that it
is \leon{beneficial} to use the density/coverage of each \leon{labeled}
training class as the constrain\leon{t} in learning the projection
from visual feature space to semantic space. We introduce a \leon{specific}
weight\leon{ing} strategy to compute $w_{z_{i}}$ in Section~\ref{subsec:Extreme-Value-Voc}.

The conventional $\epsilon-$SVR is formulated as a constrained
minimization problem, \ie, convex quadratic programming problem,
while $\epsilon-$SSVR employs quadratic smoothing \cite{zhang2004solving}
to make Eq.~(\ref{eq:SSVR}) differentiable everywhere, and thus $\epsilon-$SSVR
can be solved as an unconstrained minimization problem directly\footnote{In practice, our tentative experiments shows that the Eq.~(\ref{eq:data-embedding})
and Eq.~(\ref{eq:SSVR}) will lead to the similar results, on average;
but formulation in Eq.~(\ref{eq:SSVR}) is more stable and has lower
variance. }.

\noindent 
\textbf{Triplet Term:} Data term above only ensures that
labelled samples project close to their correct prototypes. However,
since it is doing so for many samples and over a number of classes,
it is unlikely that all the data constraints can be satisfied exactly.
Specifically, consider the following case, if $\mathbf{u}_{z_{i}}$
is in the part of the semantic space where no other entities live
(\ie, distance from $\mathbf{u}_{z_{i}}$ to any other prototype
in the embedding space is large), then projecting $\mathbf{x}_{i}$
further away from $\mathbf{u}_{z_{i}}$ is asymptomatic, \ie, will
not result in misclassification. However, if the $\mathbf{u}_{z_{i}}$
is close to other prototypes then minor error in regression may result
in misclassification.

To embed this intuition into our learning, we enforce more
discriminative constraints in the learned semantic embedding space.
Specifically, the distance of $D\left(\mathbf{x}_{i},\mathbf{u}_{z_{i}}\right)$
should not only be as small as possible, but should also be smaller
than the distance $D\left(\mathbf{x}_{i},\mathbf{u}_{a}\right)$,
$\forall a\neq z_{i}$. Formally, we define the triplet term
\begin{equation}
\mathcal{M}_{V}\left(\mathbf{x}_{i},\mathbf{u}_{z_{i}}\right)=\frac{1}{2}\sum_{a=1}^{A_{V}}\left[C+\frac{1}{2}D\left(\mathbf{x}_{i},\mathbf{u}_{z_{i}}\right)-\frac{1}{2}D\left(\mathbf{x}_{i},\mathbf{u}_{a}\right)\right]_{+}^{2},\label{eq:vocab_maximal_margin}
\end{equation}
where $a\in\mathcal{W}_{t}$ \leon{(or more precisely $a\in\mathcal{W}\setminus\mathcal{W}_{s}$)}
is selected from the open vocabulary; $C$ is the margin gap constant.
Here, $\left[\cdot\right]_{+}^{2}$ indicates the quadratically smooth
hinge loss \cite{zhang2004solving} which is convex
and has the gradient at every point. To speedup computation, we use
the closest $A_{V}$ target prototypes to each source/auxiliary prototype
$\mathbf{u}_{z_{i}}$ in the semantic space. We also define similar
constraints for the source prototype pairs: 
\begin{equation}
\mathcal{M}_{S}\left(\mathbf{x}_{i},\mathbf{u}_{z_{i}}\right)=\frac{1}{2}\sum_{b=1}^{B_{S}}\left[C+\frac{1}{2}D\left(\mathbf{x}_{i},\mathbf{u}_{z_{i}}\right)-\frac{1}{2}D\left(\mathbf{x}_{i},\mathbf{u}_{b}\right)\right]_{+}^{2}\label{eq:source_maximal_margin}
\end{equation}
where $b\in\mathcal{W}_{s}$ is selected from source/auxiliary dataset
vocabulary. This term enforces that $D\left(\mathbf{x}_{i},\mathbf{u}_{z_{i}}\right)$
should be smaller than the distance $D\left(\mathbf{x}_{i},\mathbf{u}_{b}\right)$,
$\forall b\neq z_{i}$. To facilitate the computation, we similarly
use closest $B_{S}$ prototypes that are closest to each prototype
$\mathbf{u}_{z_{i}}$ in the source classes. Note that, the Crammer
and Singer loss \cite{tsochantaridis2005large,crammer2001algorithmic}
is the upper bound of Eq.~(\ref{eq:vocab_maximal_margin}) and (\ref{eq:source_maximal_margin})
which we use to tolerate slight variants of $\mathbf{u}_{z_{i}}$
(\eg, the prototypes of 'pigs' Vs. 'pig').

To sum up, the complete triplet maximum margin term is: 
\begin{equation}
\mathcal{M}\left(\mathbf{x}_{i},\mathbf{u}_{z_{i}}\right)=\mathcal{M}_{V}\left(\mathbf{x}_{i},\mathbf{u}_{z_{i}}\right)+\mathcal{M}_{S}\left(\mathbf{x}_{i},\mathbf{u}_{z_{i}}\right).\label{eq:maximal_margin_term}
\end{equation}

\noindent We note that the form of rank hinge loss in
Eq.~(\ref{eq:vocab_maximal_margin}) and (\ref{eq:source_maximal_margin})
is similar to DeViSE \cite{frome2013devise}, but DeViSE only considers
loss with respect to source/auxiliary data and prototypes.

\noindent
\textbf{Maximum Margin Vocabulary-informed Embedding:} The complete combined objective
can now be written as: 

\begin{eqnarray}
W=\underset{W}{\mathrm{argmin}}\sum_{i=1}^{n_{T}}(\alpha\mathbb{\mathcal{L}_{\epsilon}}\left(\mathbf{x}_{i},\mathbf{u}_{z_{i}}\right)+~~~~~~~~~~~~~~~~~~~~~~~~~\nonumber \\
(1-\alpha)\mathcal{M}\left(\mathbf{x}_{i},\mathbf{u}_{z_{i}}\right))+\lambda\parallel W\parallel_{F}^{2},\label{eq:formulation}
\end{eqnarray}

\noindent where $\alpha\in[0,1]$ is the coefficient that controls contribution of the two terms.
One practical advantage is that the objective function in Eq.~(\ref{eq:formulation})
is an unconstrained minimization problem which is differentiable and
can be solved with L-BFGS. $W$ is initialized with all zeros and
converges in $10-20$ iterations.


\subsection{Weighted Maximum Margin Voc Embedding (WMM-Voc)\label{subsec:Extreme-Value-Voc}\textcolor{black}{{} }}

\begin{figure}
\begin{centering}
\includegraphics[scale=0.5]{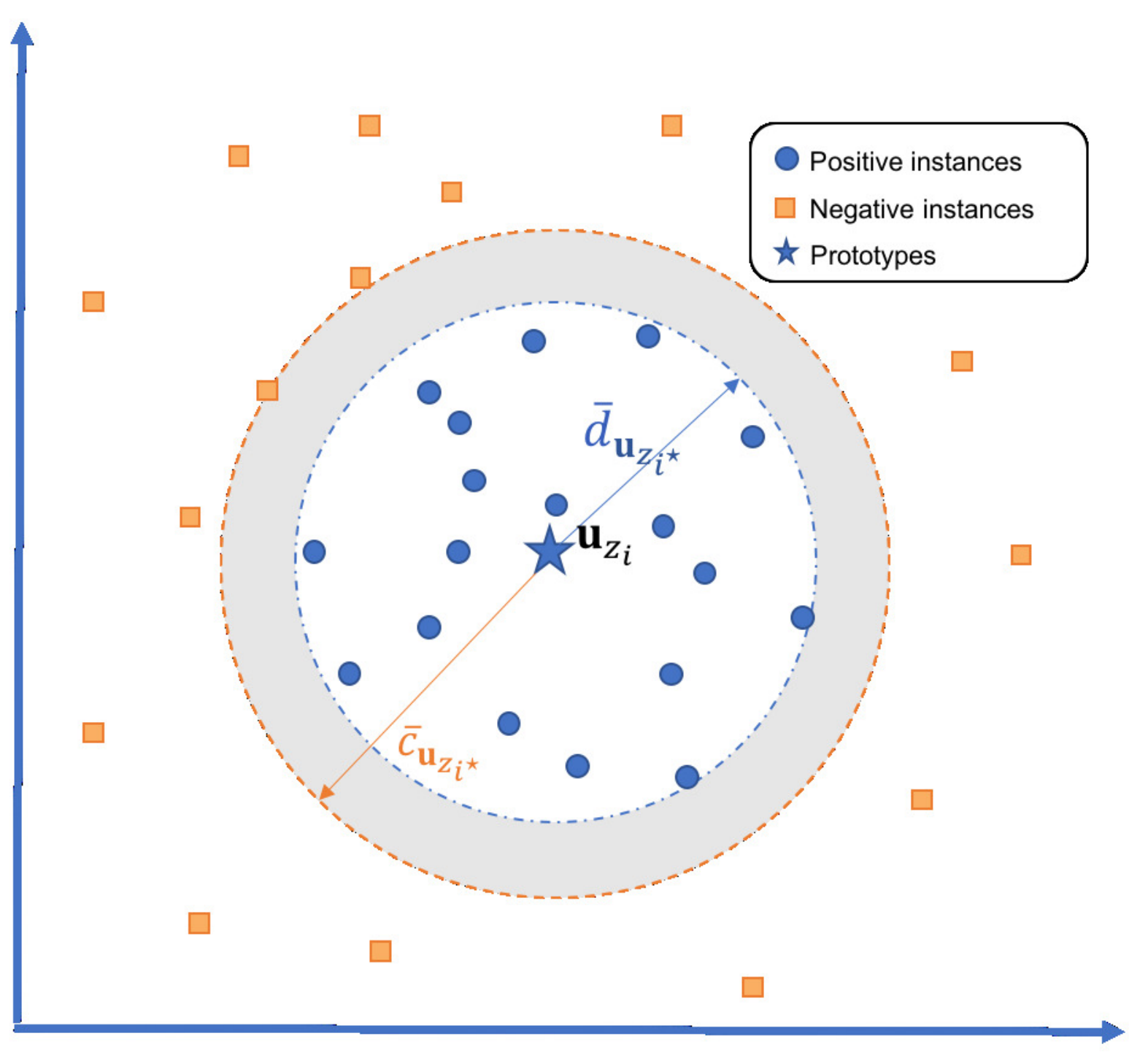} 
\par\end{centering}
\caption{\label{fig:Illustration-of-margin}Illustration of margin distribution
of prototypes in the semantic space.}
\end{figure}

We note that there is no previous method that directly estimates the
density of source training classes in the semantic space. However,
doing so may lead to several benefits. 
First, the number of training instances in source classes may be unbalanced.
In such a case, an estimate of the density of samples in a training class
can be utilized as a constraint in learning the embedding characterized by Eq.~(\ref{eq:epsilon}).
Second, in the semantic space, the instances from
the classes whose data samples span a large radius \cite{rudd2017extreme} may
reside in the neighborhood of many other classes or open vocabulary.
This can happen when the embedding is not well learned. We can interpret
this phenomenon as hubness \cite{shigeto2015ridge,lazaridou2015hubness}\footnote{However, the causes for hubness are still under investigation \cite{dinu2014improving,shigeto2015ridge}.}.
Adding a penalty based on the density of each training class may be helpful
in better learning the embedding and alleviating the hubness problem. 

This subsection introduces a strategy for estimating the density of each
known class in the semantic space (\ie, $w_{z_{i}}$ in Eq.~(\ref{eq:epsilon})).
Generally, we know the prototype of each known and novel class in
the semantic space. To estimate the density/coverage of a known
class, one  needs to look at pairwise distance between a prototype
and the nearest negative instance and the furthest positive instance. 
This intuition leads us to introduce the concept of margin distribution.


\noindent \textbf{\noindent Margin Distribution:} The concept of
margin is fundamental to maximum margin classifiers (\eg, SVMs) in
machine learning. The margin enables an intuitive interpretation of
such classifiers in searching for the maximum margin separator in
a Reproducing Kernel Hilbert Space. Previous margin classifiers \cite{zhou2014large}
aim to maximize a single margin across all training instances.
In contrast, some recent studies \cite{zhang2014large,rudd2017extreme,schapire1998boosting,dong2018learning}
suggest that the knowledge of margin distribution of instances, rather than a single
margin across all instances, is crucial for improving the generalization performance
of a classifier.

The ``instance margin'' is defined as the distance between one instance and the separating hyperplane.
Formally, for one instance $i$ in the semantic space $g\left(\mathbf{x}_{i}\right)$ and sufficiently many\footnote{In our experiments, 
we use all available training instances here.} samples $g\left(\mathbf{x}_{j}\right)$
($z_{i}\neq z_{j}$) drawn from well behaved class distributions\footnote{The well behaved indicates that the moments of the
distribution should be well-defined. For example, Cauchy distribution
is not well-behaved \cite{kotz2004continuous}.}. We define the distance
$d_{ij}=\left\Vert g\left(\mathbf{x}_{i}\right)-g\left(\mathbf{x}_{j}\right)\right\Vert $.
For instance $i$, we can obtain a set of distances $D_{i}=\left\{ d_{ij},z_{j}\neq z_{i}\right\} $
with the minimal values $\bar{d}_{i\star}=\underset{}{\mathrm{min}}{\displaystyle }D_{i}$.
As shown in \cite{rudd2017extreme}, the distribution for the minimal values
of the margin distance is characterized by a Weibull distribution. 
Based on this finding, we can express 
the probability of $g\left(\mathbf{x}\right)$
being included in the boundary estimated by $g\left(\mathbf{x}_{i}\right)$: 
\begin{equation}
\psi\left(g\left(\mathbf{x}\right);g\left(\mathbf{x}_{i}\right)\right)=\exp\left(-\left(\frac{\left\Vert g\left(\mathbf{x}\right)-g\left(\mathbf{x}_{i}\right)\right\Vert }{\lambda_{i}}\right)^{\kappa_{i}}\right),\label{eq:margin_distribution-1}
\end{equation}
\noindent where $\kappa_{i}$ and $\lambda_{i}$
are Weibull shape and scale parameters 
obtained by fitting $D_{i}$ using Maximum Likelihood Estimate (MLE), which is summarized\footnote{codes released in \url{https://github.com/xiaomeiyy/WMM-Voc}.} in Alg. \ref{alg:MLE}. 
Equation~(\ref{eq:margin_distribution-1}) quantitatively describes the margin of one 
specific class, probabilistically, in our semantic embedding space. Note that
Eq.~(\ref{eq:margin_distribution-1}) requires $\psi\left(\cdot\right)$
to be non-degenerate margin distribution, which is essentially guaranteed
by Extreme Value Theorem \cite{kotz2000extreme}. 


\begin{algorithm}[H]
\textcolor{black}{\underline{Input}: Extreme values $x_{1},\cdots,x_{n}$}

\textcolor{black}{\underline{Output}: Estimated parameters $\hat{\kappa},\hat{\lambda}$}

\textcolor{black}{If $n==1$:}

\quad{}\textcolor{black}{$\hat{\kappa}=\infty,\hat{\lambda}=x_{1}$.}

\textcolor{black}{Else:}

\quad{}\textcolor{black}{1. Sort $x_{1},\cdots,x_{n}$ to get $x_{[1]}\geq\cdots\geq x_{[n]}$}

\quad{}\quad{}\quad{}\textcolor{black}{(where $x_{[i]}$ is the re-ordered value).}

\quad{}\textcolor{black}{2. Maximum likelihood estimator for $\kappa$: 
\begin{equation}
\frac{n\sum\left(x_{[i]}^{\kappa}\log x_{[i]}-x_{[n]}^{\kappa}\log x_{[n]}\right)}{\sum\left(x_{[i]}^{\kappa}-x_{[n]}^{\kappa}\right)}=\sum\log x_{[i]}\label{eq:kappa}
\end{equation}}

\quad{}\textcolor{black}{3. Solve Eq. (\ref{eq:kappa}), and numerically estimate $\hat{\kappa}$.}

\quad{}\quad{}\quad{}\textcolor{black}{(\eg, using {\tt fzero} function in MATLAB)}

\quad{}\textcolor{black}{4. Compute $\hat{\lambda}=\left(\sum\left(x_{[i]}^{\hat{\kappa}}-x_{[n]}^{\hat{\kappa}}\right)/n\right)^{1/\hat{\kappa}}$.}

\caption{\textcolor{black}{EVT estimator by Weibull distribution}.\label{alg:MLE}}
\end{algorithm}

\noindent \textbf{\noindent Margin Distribution of Prototypes:} 
Consider a class  $z_{i}$ which in the embedding space is represented by
a prototype $\mathbf{u}_{z_{i}}$.
In accordance with above formalism, we can also assume sufficiently many 
samples $g\left(\mathbf{x}_{j}\right)$ drawn from other 
($z_{i}\neq z_{j}$) well behaved class distributions.
We can also consider the prototypes of vast open vocabulary $\mathbf{u}_{z_{j}}$
(\emph{$z_{i}\neq z_{j}$}, $z_{j}\in\mathcal{W}_{t}$). 
Under these assumptions,  we can obtain a set of distances 
$D_{\mathbf{u}_{z_{i}}}=\{\left\Vert \mathbf{u}_{z_{i}}-\mathbf{g}_{z_{j}}\right\Vert ,z_{j}\neq z_{i},\mathbf{g}_{z_{j}}\in\left\{ g\left(\mathbf{x}_{j}\right),\mathbf{u}_{z_{j}}\right\} \}$ for the prototype $\mathbf{u}_{z_{i}}$.
As a result, the distribution for the minimal values of the margin distance
for $\mathbf{u}_{z_{i}}$ is given by a Weibull distribution. 
The probability that $\mathbf{g}_{z_{i}}$ is included in the boundary estimated
by $\mathbf{u}_{z_{i}}$ is given by
\begin{equation}
\psi\left(\mathbf{g}_{z_{i}};\mathbf{u}_{z_{i}}\right)=\exp\left(-\left(\frac{\left\Vert \mathbf{g}_{z_{i}}-\mathbf{u}_{z_{i}}\right\Vert }{\lambda_{\mathbf{u}_{z_{i}}}}\right)^{\kappa_{\mathbf{u}_{z_{i}}}}\right).\label{eq:margin-dist-1}
\end{equation}
%
\noindent The above equation models the distribution of minimum value;
thus it can be used to estimate the boundary density (or more specifically,
the boundary distribution) of class $z_{i}$.

We set significant level to 0.05 to approximately
estimate the minimal value $\bar{d}_{\mathbf{u}_{z_{i}}\star}$.
As illustrated in Figure \ref{fig:Illustration-of-margin},
if $\psi\left(\mathbf{g}_{z_{i}};\mathbf{u}_{z_{i}}\right)<0.05,$ we will
assume $\mathbf{g}_{z_{i}}$ does not belong to the prototype $\mathbf{u}_{z_{i}}$;
otherwise, $\mathbf{g}_{z_{i}}$ is included in the boundary estimated by $\mathbf{u}_{z_{i}}$.
In term of the significant level of 0.05, we can further denote the minimal
values as $\bar{d}_{\mathbf{u}_{z_{i}}\star}^{\left(0.05\right)}$,
\ie, $\exp\left(-\left(\frac{\bar{d}_{\mathbf{u}_{z_{i}}\star}^{\left(0.05\right)}}{\lambda_{\mathbf{u}_{z_{i}}}}\right)^{\kappa_{\mathbf{u}_{z_{i}}}}\right)=0.05$.
Thus we have

\begin{equation}
\bar{d}_{\mathbf{u}_{z_{i}}\star}^{\left(0.05\right)}=\lambda_{\mathbf{u}_{z_{i}}}\cdot \log^{1/\kappa_{\mathbf{u}_{z_{i}}}}\left(\frac{1}{0.05}\right)\label{eq:positive}
\end{equation}

\noindent \textbf{\noindent Coverage Distribution of Prototypes:}
Now, for class $z_{i}$ consider the nearest instance from another class
$g\left(\mathbf{x}_{j}\right)$ where $z_{i}\neq z_{j}$; with sufficient
many instances $g\left(\mathbf{x}_{k}\right)$ from class $z_{i}$,
we have pairwise unique ("within class") distance:
\begin{equation}
c_{\mathbf{u}_{z_{i}}k}=\Vert g\left(\mathbf{x}_{k}\right)-\mathbf{u}_{z_{i}}\Vert.\label{eq:coverage-inequality}
\end{equation}
We consider outliers those instances $g\left(\mathbf{x}_{k}\right)$
 that have larger distance to $\mathbf{u}_{z_{i}}$
than the nearest instance $g\left(\mathbf{x}_{j}\right)$ ($z_{i}\neq z_{j}$) of another class.
To remove the outliers we hence consider 
$C_{\mathbf{u}_{z_{i}}}=\left\{ c_{\mathbf{u}_{z_{i}}k}|c_{\mathbf{u}_{z_{i}}k}\leq\min_{{z}_{j}\neq{z}_{k}}\Vert g\left(\mathbf{x}_{j}\right)-\mathbf{u}_{z_{i}}\Vert\right\} $. 
As illustrated in Figure \ref{fig:Illustration-of-margin}, we only consider positive instances within the orange circle and all other instances with larger distance are removed.
Then the distribution of the largest distance $\bar{c}_{\mathbf{u}_{z_{i}}\star}=\underset{}{\max C_{\mathbf{u}_{z_{i}}}}$
will follow a reversed Weibull distribution. This allows us to get the
probability distribution to describe positive instances, 
\begin{equation}
\phi\left(g\left(\mathbf{x}_{k}\right);\mathbf{u}_{z_{i}}\right)=1-\exp\left(-\left(\frac{\left\Vert g\left(\mathbf{x}_{k}\right)-\mathbf{u}_{z_{i}}\right\Vert }{\lambda_{\mathbf{u}_{z_{i}}}'}\right)^{\kappa_{\mathbf{u}_{z_{i}}}^{'}}\right)\label{eq:coverage}
\end{equation}

\noindent where $\kappa_{i}^{'}$ and $\lambda_{i}^{'}$ are reverse
Weibull shape and scale parameters individually obtained from fitting
the largest $C_{\mathbf{u}_{z_{i}}}$, $\bar{c}_{\mathbf{u}_{z_{i}}\star}$
is the distance between instance and prototype, $\phi$ is the probability
that the instance is in the class.

Similar to the margin distribution, we can estimate the
coverage by setting the significant level to $0.05$. As
shown in Figure \ref{fig:Illustration-of-margin}, we establish two boundaries to estimate the 
scale of each class probabilistically.
If $\phi\left(g\left(\mathbf{x}_{k}\right);\mathbf{u}_{z_{i}}\right)\geqslant0.05$,
$g\left(\mathbf{x}_{k}\right)$ is included in the coverage distribution
$\mathbf{u}_{z_{i}}$. The maximum values $\bar{c}_{\mathbf{u}_{z_{i}}\star}^{\left(0.05\right)}$
can be computed as $\phi\left(g\left(\mathbf{x}_{k}\right);\mathbf{u}_{z_{i}}\right)=0.05$.
It results in,

\begin{equation}
\bar{c}_{\mathbf{u}_{z_{i}}\star}^{\left(0.05\right)}=\lambda_{\mathbf{u}_{z_{i}}}'\cdot \log^{1/\kappa_{\mathbf{u}_{z_{i}}}^{'}}\left(\frac{1}{1-0.05}\right).\label{eq:negative}
\end{equation}

By combining the terms computed in Eq.~(\ref{eq:positive}) and (\ref{eq:negative}),
we can obtain the weight $w_{z_{i}}$ for class $z_{i}$ in Eq.~(\ref{eq:epsilon}),

\begin{equation}
w_{z_{i}}\propto\left(\bar{d}_{\mathbf{u}_{z_{i}}\star}^{(0.05)}+\bar{c}_{\mathbf{u}_{z_{i}}\star}^{(0.05)}\right)\label{eq:weights_combined}
\end{equation}


\textcolor{black}{As explained in Algorithm \ref{alg:MLE}, we set $\hat{\kappa}=\infty,\hat{\lambda}=x_{1}$  in one-shot setting. 
In few-shot learning setting, we can estimate $\hat{\kappa}$ and
$\hat{\lambda}$ directly. In addition, 
such an initialization of weights ($\hat{\kappa}$ and $\hat{\lambda}$) intrinsically helps learn the embedding weight $W$. }

\noindent
\textcolor{black}{\noindent \textbf{The learning process of parameters:}  The process could be interpreted as a form of block 
coordinate descent where we estimate the 
embedding/mapping; then density within that embedding and so on. 
In practice, the weights $w_{z_{i}}$ are initially randomized. But they do not play
an important role at the beginning of the optimization,
since $\left|W_{\star j}^{T}\mathbf{x}_{i}-\left(\mathbf{\mathbf{u}}_{z_{i}}\right)_{j}\right|${}
is very large in the first few iterations. In other words, the optimization is initially
dominated by the data term and maximum margin terms play little role. 
However, once we can get a relative good mapping (\ie, smaller $\left|W_{\star j}^{T}\mathbf{x}_{i}-\left(\mathbf{\mathbf{u}}_{z_{i}}\right)_{j}\right|$)
after several training iterations, the weight $w_{z_{i}}$ starts becoming
significant. 
By virtue of such an optimization, the weighted version can achieve better performance than
the previous non-weighted version in our conference paper \cite{fu2016semi}.
}

\vspace{0.05in}
\noindent \textbf{\noindent Deep Weighted Maximum Margin Voc Embedding
(Deep WMM-Voc). }In practice, we 
extend WMM-Voc to include a deep network for feature extraction. 
Rather than extracting low-level features using an off-the-shelf pre-trained model in Eq.~(\ref{eq:formulation}),
we use an integrated deep network to extract $\mathbf{x}_{i}$ from the raw images.
As a result, the loss function in Eq.~(\ref{eq:formulation}) is also used to optimize
the parameters of the deep network. In particular, we fix the convolutional
layers of corresponding network and fine-tune the \textcolor{black}{last} fully connected
layer in our task. The network was trained using stochastic
gradient descendent. 

\section{Experiments}

\subsection{Experimental setup}

\noindent We conduct our experiments on Animals with Attributes (AwA)
dataset, and ImageNet $2012$/$2010$ dataset. 


\noindent \textbf{AwA dataset:} \leon{AwA} consists of 50 classes
of animals ($30,475$ images in total). In \cite{lampert2014attribute}
standard split into 40 source/auxiliary classes ($|\mathcal{W}_{s}|=40$)
and 10 target/test classes ($|\mathcal{W}_{t}|=10$) is introduced.
We follow this split for supervised and zero-shot learning. We use
ResNet101 features (downloaded from \cite{palatucci2009zero}) on AwA to make
the results more easily comparable to state-of-the-art.


\noindent \textbf{ImageNet $2012$/$2010$ dataset:} \leon{ImageNet}
is a large-scale dataset. We use $1000$ ($|\mathcal{W}_{s}|=1000$)
classes of ILSVRC $2012$ as the source/auxiliary classes and $360$
($|\mathcal{W}_{t}|=360$) classes of ILSVRC 2010 that are not used
in ILSVRC $2012$ as target data. We use pre-trained VGG-19 model
\cite{chatfield2014return} to extract deep features for ImageNet.


\noindent \textbf{Recognition tasks:} We consider several different
settings in a variety of experiments. We first divide the two datasets
into source and target splits. On source classes, we can validate
whether our framework can be used to solve one-shot and supervised
recognition. By using both the source and target classes, transfer
learning based settings can be evaluated. 
\begin{enumerate}
\item \textbf{\textsc{Supervised}}\textbf{ recognition}: learning is on
source classes; 
test instances come from the same classes with $\mathcal{W}_{s}$
as recognition vocabulary. In particular, under this setting, we also
validate the one- \leon{and few-}shot recognition scenario\leon{s},
\emph{i.e.}, class\leon{es have} one or few training example\leon{s.} 
\item \textbf{\textsc{Zero-shot}}\textbf{ recognition}: \leon{In ZSL, learning}
is on \leon{the} source classes \leon{with $\mathcal{W}_{s}$ vocabulary;}
test instances \leon{come} from target dataset with $\mathcal{W}_{t}$
as recognition vocabulary. 
\item \textbf{\textsc{General-zero-shot}}\textbf{ recognition}: G-ZSL \leon{uses
source classes to learn, with test instances coming from either target
$\mathcal{W}_{t}$ or original $\mathcal{W}_{s}$ recognition vocabulary.} 
\item \textbf{\textsc{Open-set}}\textbf{ recognition}: \leon{Again source
classes are used for learning, but the entire open vocabulary with
$|\mathcal{W}|\approx310K$ atoms is used at test time. In practice,
test images come from both source and target splits (similar to G-ZSL);
however, unlike G-ZSL there are additional distractor classes. In
other words, chance performance for open-set recognition is much lower
than for G-ZSL.} 
\end{enumerate}
We test both our Voc \leon{variants} -- MM-Voc and WMM-Voc. Additionally,
we also validate the Deep WMM-Voc by fine-tuning the \leon{W}MM-Voc
on \textcolor{black}{VGG-19} architecture and optimizing the weights
\leon{with respect to the loss in} Eq.~(\ref{eq:formulation}).


\noindent \textbf{Competitors:} \leon{We compare to a variety of the
models in the literature, including:} 
\begin{enumerate}
\item \textbf{SVM}: SVM classifier trained directly on the training instances
of source data, without the use of semantic embedding. This is the
standard (\textsc{Supervised}/\textsc{One-shot}) learning setting
and the learned classifier can only predict the labels within the source classes. 
\leon{Hence, SVM is inapplicable in ZSL, G-ZSL, and open-set recognition settings.} 
\item \textbf{SVR-Map}: SVR is used to learn $W$ and the recognition is
done\leon{, similar to our method,} in the resulting semantic manifold.
This corresponds to only \leon{optimizing} Eq.~(\ref{eq:SSVR}). 
\item \textbf{Deep-SVR}: \leon{This is a variant SVR, which further allows
fine-tuning of the underlying neural network generating the features.
In this case, $W$ is expressed as the last linear layer and the entire
network is fine-tuned with respect to the loss encoding only the data
term (Eq.~(\ref{eq:SSVR})).} 
\item \textbf{SAE}: SAE is a semantic encoder-decoder paradigm that projects
visual features into a semantic space and then reconstructs the original
visual feature representation \cite{kodirov2017semantic}. The SAE
has two variants in learning the embedding space, \emph{i.e.}, semantic
space to feature space (S$\rightarrow$F), and feature space to semantic
space (F$\rightarrow$S). By default, the best result of these two
variants are reported. 
\item \textbf{ESZSL}: \leon{ESZSL} first learns the mapping between
visual features and attributes, then models the relationship between
attributes and classes \cite{romera2015embarrassingly}. 
\item \textbf{DeVise, ConSE, AMP}: To compare with state-of-the-art large-scale
zero-shot learning approaches we implement DeViSE \cite{frome2013devise}
and ConSE \cite{norouzi2013zero}\footnote{Codes for \cite{frome2013devise} and \cite{norouzi2013zero}
are not publicly available.}. ConSE uses a multi-class logistic regression classifier for predicting
class probabilities of source instances; and the parameter T (number
of top-T nearest embeddings for a given instance) was selected from
$\{1,10,100,1000\}$ that gives the best results. ConSE method in
supervised setting works the same as SVR. We use the AMP code provided
on the author webpage \cite{fu2015zero}. 
\end{enumerate}

\noindent \textbf{\textcolor{black}{Metrics:}}\textcolor{black}{{} Classification
accuracies are reported as the evaluation metrics on most of tasks.
In our conference version \cite{fu2016semi}, we further introduce an evaluation
setting for }\textsc{\textcolor{black}{Open-set}}\textcolor{black}{{}
tasks where we do not assume that test data comes from either source/auxiliary
domain or target domain. Thus we split the two cases (}\textcolor{black}{\emph{i.e.}}\textcolor{black}{,
}\textsc{\textcolor{black}{Supervised}}\textcolor{black}{-like, and
}\textsc{\textcolor{black}{Zero-shot}}\textcolor{black}{-like settings),
to mimic }\textsc{\textcolor{black}{Supervised}}\textcolor{black}{{} and
}\textsc{\textcolor{black}{Zero-shot}}\textcolor{black}{{} scenarios for
easier analysis. Particularly, in G-ZSL task, this newly introduced
evaluation setting is corresponding to the evaluation metrics defined
in \cite{xian2017zero}: (1) $\mathbb{S}\rightarrow\mathbb{T}$: Test
instances from seen classes, the prediction candidates include both
seen and unseen classes; (2) $\mathbb{U}\rightarrow\mathbb{T}$: Test
instances from unseen classes, the prediction candidates include both
seen and unseen classes. (3) The harmonic mean is used as the main
evaluation metric to further combine the results of both $\mathbb{S}\rightarrow\mathbb{T}$
and $\mathbb{U}\rightarrow\mathbb{T}$:
\begin{equation}
H=2\text{\ensuremath{\cdot}}\frac{\left(Acc(\mathbb{U}\rightarrow\mathbb{T})\times Acc(\mathbb{S}\rightarrow\mathbb{T})\right)}{\left(Acc(\mathbb{U}\rightarrow\mathbb{T})+Acc(\mathbb{S}\rightarrow\mathbb{T})\right)}.\label{eq:harmonic}
\end{equation}
}


\noindent \textbf{Setting of Parameters:}  \textcolor{black}{For
the recognition tasks, we learn classifiers by using various number
of training instances. }We compare \leon{relevant baselines with
results of our method variants: MM-Voc, WMM-Voc, Deep WMM-Voc.} Each
setting is repeated/tested 10 times. The averaged results are reported
to reduce the variance. For each setting, our Voc methods \leon{are}
trained by a single model to be capable of solving the tasks of supervised,
zero-shot, G-ZSL \leon{and open-set} recognition. Specifically, 
\begin{enumerate}
\item In Deep WMM-Voc, we fix $\lambda$ to $0.01$ and $\alpha=0.6$ with
the learning rate initially set to $1e-5$ and is reduced by {\small $\frac{1}{2}$} every
$10$ epochs. $A_{V}$ and $B_{S}$ are set to $5$ in order to balance
performance and computational cost of pairwise constraints. 
\item To solve Eq.~(\ref{eq:formulation}) at a scale, one can use Stochastic
Gradient Descent (SGD) which makes great progress initially, but often
is slow when approaching convergence. In contrast, the L-BFGS method
mentioned above can achieve steady convergence at the cost of computing
the full objective and gradient at each iteration. L-BFGS can usually
achieve better results than SGD with good initialization, however,
is computationally expensive. To leverage benefits of both of these
methods, we utilize a hybrid method to solve Eq.~(\ref{eq:formulation})
in large-scale datasets: the solver is initialized with few instances
to approximate the gradients using SGD first; then gradually more
instances are used and switch to L-BFGS is made with iterations. This
solver is motivated by Friedlander \etal~\cite{friedlander2012hybrid},
who theoretically analyzed and proved the convergence for the hybrid
optimization methods. In practice, we use L-BFGS and the Hybrid algorithms
for AwA and ImageNet respectively. The hybrid algorithm can save between
$20\sim50\%$ training time as compared with L-BFGS. 
\end{enumerate}

\noindent \textbf{Open set vocabulary.} We use Google word2vec to
learn the open set vocabulary set from a large text corpus of around
$7$ billion words: UMBC WebBase ($3$ billion words), the latest
Wikipedia articles ($3$ billion words) and other web documents ($1$
billion words). Some rare (low frequency) words and high frequency
stopping words were pruned in the vocabulary set: we remove words
with the frequency $<300$ or $>10\ million$ times. The result is
a vocabulary of around 310K words/phrases with $openness\approx1$,
which is defined as $openness=1-\sqrt{\left(2\times|\mathcal{W}_{s}|\right)/\left(|\mathcal{W}|\right)}$
\cite{Scheirer_2013_TPAMI}.

\subsection{Experimental results on AwA dataset}

\begin{figure}
\begin{centering}
\includegraphics[scale=0.37]{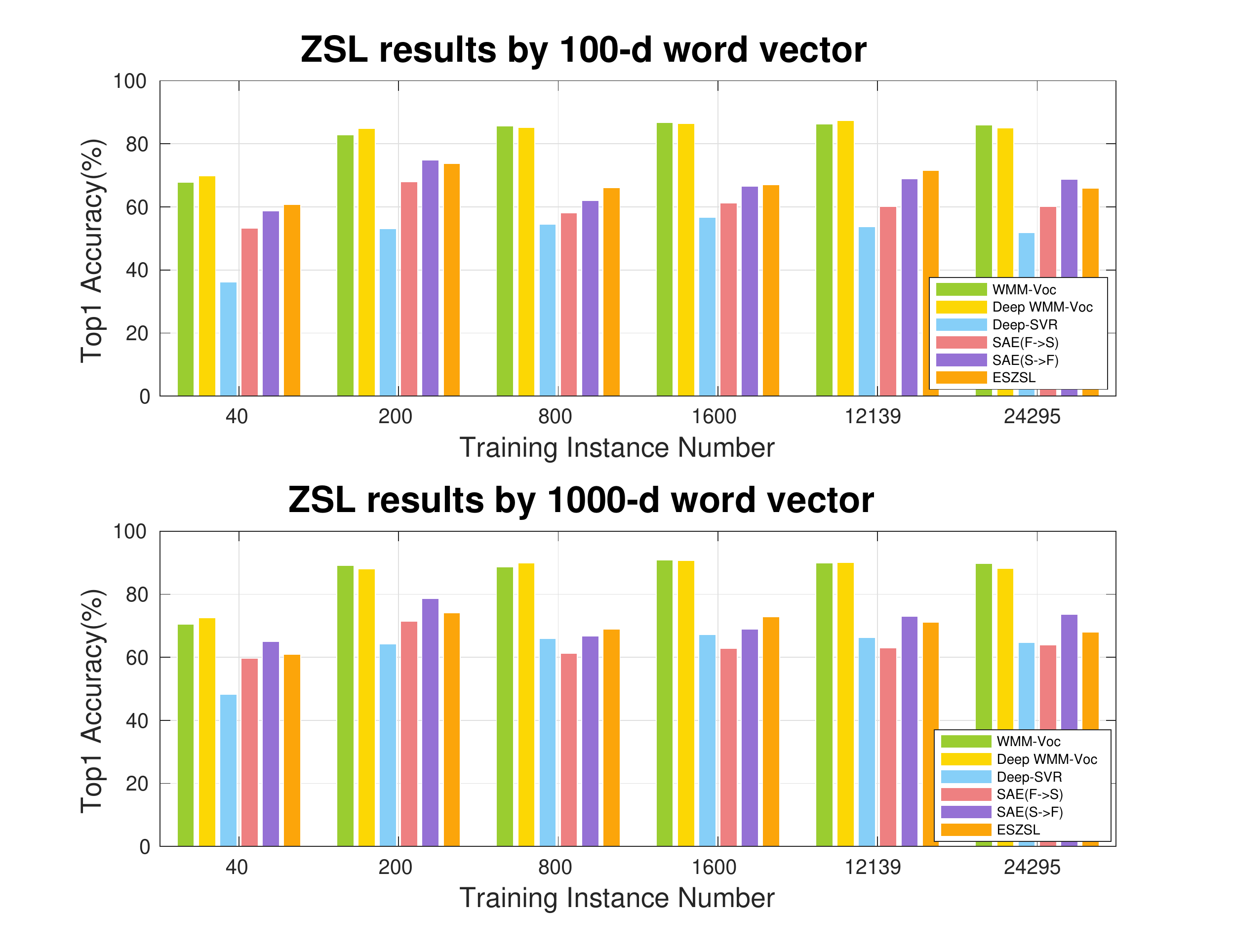} 
\par\end{centering}
\caption{\label{fig:AwA-ZSL-results.}The ZSL results on AwA by different settings.}
\end{figure}

\noindent 
\begin{table}
\begin{centering}
{\small{}}%
\begin{tabular}{c|c|c|c}
\hline 
{\small{}Methods } & {\small{}S. Sp } & {\small{}Features } & {\small{}Acc. }\tabularnewline
\hline 
\hline 
\textbf{\small{}WMM-Voc} & {\small{}{}W }  & {\small{}{}$CNN_{\text{\tiny resnet101}}$ }  & {\small{}$\textbf{90.79}$}   \tabularnewline
\textbf{\small{}WMM-Voc: closed} & {\small{}{}W }  & {\small{}{}$CNN_{\text{\tiny resnet101}}$ }  & {\small{}84.51}   \tabularnewline
\hline 
\textbf{\small{}{}Deep WMM-Voc}{\small{}{} }  & {\small{}{}W }  & {\small{}{}$CNN_{\text{\tiny resnet101}}$ }  & {\small{}$\textbf{90.65}$}\tabularnewline
\textbf{\small{}Deep WMM-Voc: closed} & {\small{}{}W }  & {\small{}{}$CNN_{\text{\tiny resnet101}}$ }  & {\small{}83.85}\tabularnewline
\hline 
\hline 
{\small{}SAE} & {\small{}W } & {\small{}$CNN_{\text{\tiny resnet101}}$ } & {\small{}$71.42$}\tabularnewline
\hline 
{\small{}ESZSL } & {\small{}W } & {\small{}$CNN_{\text{\tiny resnet101}}$ } & {\small{}$74.17$}\tabularnewline
\hline 
{\small{}Deep-SVR } & {\small{}W } & {\small{}$CNN_{\text{\tiny resnet101}}$ } & {\small{}$67.22$}\tabularnewline
\hline 
\hline 
{\small{}Akata }\emph{\small{}et al.}{\small{} \cite{akata2015evaluation} } & {\small{}A+W } & {\small{}$CNN_{\text{\tiny GoogleNet}}$ } & {\small{}$73.90$}\tabularnewline
\hline 
{\small{}TMV-BLP \cite{fu2014transductive} } & {\small{}A+W } & {\small{}$CNN_{\text{\tiny OverFeat}}$ } & {\small{}$69.90$}\tabularnewline
\hline 
{\small{}AMP (SR+SE) \cite{fu2015zero} } & {\small{}A+W } & {\small{}$CNN_{\text{\tiny OverFeat}}$ } & {\small{}$66.00$}\tabularnewline
\hline 
{\small{}PST\cite{rohrbach2013transfer} } & {\small{}A+W } & {\small{}$CNN_{\text{\tiny OverFeat}}$ } & {\small{}$54.10$}\tabularnewline
\hline 
{\small{}Latem \cite{xian2016latent} } & {\small{}A+W } & {\small{}$CNN_{\text{\tiny resnet101}}$ } & {\small{}74.80 }\tabularnewline
\hline 
{\small{}SJE \cite{akata2015evaluation} } & {\small{}A+W } & {\small{}$CNN_{\text{\tiny resnet101}}$ } & {\small{}76.70 }\tabularnewline
\hline 
\hline 

{\small{}DeViSE \cite{frome2013devise} } & {\small{}W} & {\small{}$CNN_{\text{\tiny resnet101}}$ } & {\small{}72.90 }\tabularnewline
\hline 
{\small{}ConSE \cite{norouzi2013zero} } & {\small{}W} & {\small{}$CNN_{\text{\tiny resnet101}}$ } & {\small{}63.60 }\tabularnewline
\hline 
{\small{}CMT \cite{socher2013zero} } & {\small{}W} & {\small{}$CNN_{\text{\tiny resnet101}}$ } & {\small{}58.90 }\tabularnewline
\hline 
{\small{}SSE \cite{zhang2015zero}} & {\small{}W} & {\small{}$CNN_{\text{\tiny resnet101}}$ } & {\small{}54.50 }\tabularnewline
\hline 
\textcolor{black}{\small{}SSE \cite{zhang2015zero}} & \textcolor{black}{\small{}W} & \textcolor{black}{\small{}$CNN_{\text{\tiny VGG19}}$ } & \textcolor{black}{\small{}57.49}\tabularnewline
\hline 
\textcolor{black}{\small{}TASTE\cite{yu2018transductive}} & \textcolor{black}{\small{}W} & \textcolor{black}{\small{}$CNN_{\text{\tiny VGG19}}$ } & \textcolor{black}{\small{}89.40}\tabularnewline
\hline 
\textcolor{black}{\small{}KLDA+KRR\cite{long2018zero}} & \textcolor{black}{\small{}W} & \textcolor{black}{\small{}$CNN_{\text{\tiny GoogleNet}}$ } & \textcolor{black}{\small{}79.30}\tabularnewline
\hline 
\textcolor{black}{\small{}CLN+KRR\cite{long2018zero}} & \textcolor{black}{\small{}W} & \textcolor{black}{\small{}$CNN_{\text{\tiny VGG19}}$ } & \textcolor{black}{\small{}81.00}\tabularnewline
\hline 
\textcolor{black}{\small{}UVDS\cite{long2018zeroSyn}} & \textcolor{black}{\small{}W} & \textcolor{black}{\small{}$CNN_{\text{\tiny VGG19}}$ } & \textcolor{black}{\small{}62.88}\tabularnewline

\hline 
\textcolor{black}{\small{}DEM\cite{changpinyo2017predicting}} & \textcolor{black}{\small{}W} & \textcolor{black}{\small{}$CNN_{\text{\tiny Inception-V2}}$ } & \textcolor{black}{\small{}86.70}\tabularnewline
\hline 
{\small{}DS \cite{rohrbach2010helps} } & {\small{}W/A} & {\small{}$CNN_{\text{\tiny OverFeat}}$ } & {\small{}$52.70$ }\tabularnewline
\hline 
{\small{}SYNC \cite{changpinyo2016synthesized} } & \textcolor{black}{\small{}W/A} & {\small{}$CNN_{\text{\tiny resnet101}}$ } & {\small{}72.20 }\tabularnewline
\hline 
\hline 
\textcolor{black}{\small{}Relation Net\cite{sung2018learning}} & \textcolor{black}{\small{}A} & \textcolor{black}{\small{}$CNN_{\text{\tiny Inception-V2}}$ } & \textcolor{black}{\small{}84.50}\tabularnewline
\hline 

{\small{}ESZSL \cite{romera2015embarrassingly} } & \textcolor{black}{\small{}A} & {\small{}$CNN_{\text{\tiny resnet101}}$ } & {\small{}74.70 }\tabularnewline
\hline 
\textcolor{black}{\small{}UVDS\cite{long2018zeroSyn}} & \textcolor{black}{\small{}A} & \textcolor{black}{\small{}$CNN_{\text{\tiny GoogleNet}}$ } & \textcolor{black}{\small{}80.28}\tabularnewline
\hline 
\textcolor{black}{\small{}GFZSL\cite{verma2017simple}} & \textcolor{black}{\small{}A} & \textcolor{black}{\small{}$CNN_{\text{\tiny VGG19}}$ } & \textcolor{black}{\small{}80.50}\tabularnewline
\hline 
\textcolor{black}{\small{}DEM\cite{changpinyo2017predicting}} & \textcolor{black}{\small{}A} & \textcolor{black}{\small{}$CNN_{\text{\tiny Inception-V2}}$ } & \textcolor{black}{\small{}78.80}\tabularnewline
\hline 
\textcolor{black}{\small{}SE-GZSL\cite{verma2018generalized}} & \textcolor{black}{\small{}A} & \textcolor{black}{\small{}$CNN_{\text{\tiny VGG19}}$ } & \textcolor{black}{\small{}69.50}\tabularnewline
\hline 
\textit{\textcolor{black}{\small{}cycle-}}\textcolor{black}{\small{}CLSWGAN\cite{felix2018multi}} & \textcolor{black}{\small{}A} & \textcolor{black}{\small{}$CNN_{\text{\tiny resnet101}}$ } & \textcolor{black}{\small{}66.30}\tabularnewline
\hline 
\textit{\textcolor{black}{\small{}f}}\textcolor{black}{\small{}-CLSWGAN\cite{xian2018feature}} & \textcolor{black}{\small{}A} & \textcolor{black}{\small{}$CNN_{\text{\tiny resnet101}}$ } & \textcolor{black}{\small{}68.20}\tabularnewline
\hline 
\textcolor{black}{\small{}PTMCA\cite{long2018pseudo}} & \textcolor{black}{\small{}A} & \textcolor{black}{\small{}$CNN_{\text{\tiny resnet101}}$ } & \textcolor{black}{\small{}66.20}\tabularnewline
\hline 
{\small{}Jayaraman }\emph{\small{}et al.}{\small{} \cite{jayaraman2014zero} } & {\small{}A} & {\small{}low-level } & {\small{}$48.70$ }\tabularnewline
\hline 
{\small{}DAP \cite{lampert2014attribute} } & {\small{}A} & {\small{}$CNN_{\text{\tiny VGG19}}$ } & {\small{}$57.50$}\tabularnewline
\hline 
{\small{}DAP \cite{lampert2014attribute} } & {\small{}A} & {\small{}$CNN_{\text{\tiny resnet101}}$ } & {\small{}57.10 }\tabularnewline
\hline 
{\small{}DAP \cite{lampert2014attribute} } & {\small{}A} & {\small{}$CNN_{\text{\tiny OverFeat}}$ } & {\small{}$53.20$}\tabularnewline
\hline 
{\small{}ALE \cite{akata2015label} } & {\small{}A} & {\small{}$CNN_{\text{\tiny resnet101}}$ } & {\small{}78.60 }\tabularnewline
\hline 
{\small{}Yu }\emph{\small{}et al.}{\small{} \cite{yu2013designing} } & {\small{}A} & {\small{}low-level } & {\small{}$48.30$}\tabularnewline
\hline 
{\small{}IAP \cite{lampert2014attribute} } & {\small{}A} & {\small{}$CNN_{\text{\tiny OverFeat}}$ } & {\small{}$44.50$}\tabularnewline
\hline 
{\small{}HEX \cite{deng2014large} } & {\small{}A} & {\small{}$CNN_{\text{\tiny DECAF}}$ } & {\small{}$44.20$}\tabularnewline
\hline 
{\small{}AHLE \cite{akata2015label} } & {\small{}A} & {\small{}low-level } & {\small{}$43.50$}\tabularnewline
\hline 
\end{tabular}{\small\par}
\par\end{centering}
\caption{\label{tab:Zero-shot-Learning-Comparison} \textbf{Zero-shot comparison
on AwA.} We compare the state-of-the-art ZSL results using different
semantic spaces (S. Sp) including word vector (W) and attribute (A).
1000 dimension word2vec dictionary is used for our model. (Chance-level
=$10\%$). Different types of  features
are used by different methods. \textcolor{black}{ \textbf{WMM-Voc: closed} and  \textbf{Deep WMM-Voc: closed }are the two variants of our model obtained by learning the vocabulary-informed constraints only from known classes (\ie, closed set), similar to our conference version \cite{fu2016semi}. }}
 
\end{table}

\subsubsection{Learning Classifiers from Few Source Training Instances}

\begin{table*}
\begin{centering}
\textcolor{black}{\small{}}%
\begin{tabular}{l|c|c|c|c|c||c|c|c}
\hline 
 & \textcolor{black}{\small{}Dimension } & \textcolor{black}{\small{}SVR-Map } & \textcolor{black}{\small{}Deep-SVR } & \textcolor{black}{\small{}SAE } & \textcolor{black}{\small{}ESZSL } & \textbf{\textcolor{black}{\small{}MM-Voc}}\textcolor{black}{\small{} } & \textbf{\textcolor{black}{\small{}WMM-Voc}}\textcolor{black}{\small{} } & \textbf{\textcolor{black}{\small{}Deep WMM-Voc}}\textcolor{black}{\small{} }\tabularnewline
\hline 
\hline 
\multirow{2}{*}{\textcolor{black}{\small{}Supervised }} & \textcolor{black}{\small{}100-dim } & \textcolor{black}{\small{}51.4/- } & \textcolor{black}{\small{}71.59/91.98 } & \textcolor{black}{\small{}70.22/92.60 } & \textcolor{black}{\small{}74.86/94.85} & \textcolor{black}{\small{}58.01/87.88 } & \textcolor{black}{\small{}75.57/94.31 } & \textcolor{black}{\small{}76.23/94.85}\tabularnewline
 & \textcolor{black}{\small{}1000-dim } & \textcolor{black}{\small{}57.1/- } & \textcolor{black}{\small{}76.32/95.22 } & \textcolor{black}{\small{}75.32/94.17 } & \textcolor{black}{\small{}75.08/94.27} & \textcolor{black}{\small{}59.1/77.73 } & \textbf{\textcolor{black}{\small{}79.44}}\textcolor{black}{\small{}/96.01 } & \textcolor{black}{\small{}76.55/{\bf 96.22}}\tabularnewline
\hline 
\multirow{2}{*}{\textcolor{black}{\small{}Zero-shot }} & \textcolor{black}{\small{}100-dim } & \textcolor{black}{\small{}52.1/- } & \textcolor{black}{\small{}53.12/84.24 } & \textcolor{black}{\small{}67.96/95.08 } & \textcolor{black}{\small{}73.69/95.83} & \textcolor{black}{\small{}61.10/96.02 } & \textcolor{black}{\small{}82.78/98.92 } & \textcolor{black}{\small{}84.87/98.87}\tabularnewline
 & \textcolor{black}{\small{}1000-dim } & \textcolor{black}{\small{}58.0/- } & \textcolor{black}{\small{}64.29/88.71 } & \textcolor{black}{\small{}71.42/97.18 } & \textcolor{black}{\small{}74.17/97.12} & \textcolor{black}{\small{}83.84/96.74  } & \textbf{\textcolor{black}{\small{}89.09/}}\textcolor{black}{\small{}99.21 } & \textcolor{black}{\small{}88.07/{\bf 99.40}}\tabularnewline
\hline 
\multirow{2}{*}{\textcolor{black}{\small{}G-ZSL }} & \textcolor{black}{\small{}100-dim } & \textcolor{black}{\small{}- } & \textcolor{black}{\small{}5.65/54.45 } & \textcolor{black}{\small{}2.15/52.7 } & \textcolor{black}{\small{}2.88/68.37} & \textcolor{black}{\small{}19.74/85.79 } & \textcolor{black}{\small{}28.92/88.01 } & \textcolor{black}{\small{}33.04/89.11}\tabularnewline
 & \textcolor{black}{\small{}1000-dim } & \textcolor{black}{\small{}- } & \textcolor{black}{\small{}0/39.84 } & \textcolor{black}{\small{}0/35.91 } & \textcolor{black}{\small{}0/33.09} & \textcolor{black}{\small{}8.54/59.79 } & \textcolor{black}{\small{}27.98/90.47 } & \textbf{\textcolor{black}{\small{}34.77/90.76}}\tabularnewline
\hline 
\end{tabular}{\small\par}
\par\end{centering}
\caption{\textcolor{black}{\small{}\label{tab:Standard-split-on}Classification
accuracy (Top-1 / Top-5) on AwA dataset for }\textsc{\textcolor{black}{\small{}Supervised}}\textcolor{black}{\small{},}\textsc{\textcolor{black}{\small{}
General zero-shot}}\textcolor{black}{\small{} and }\textsc{\textcolor{black}{\small{}Zero-shot}}\textcolor{black}{\small{}
settings for 100-dim and 1000-dim word2vec representation (200 instances).}}
\end{table*}

\textcolor{black}{We are particularly interested in learning of classifiers
from few source training instances. This is inclined to mimic
human performance of learning from few examples and illustrate ability
of our model to learn with little data}\footnote{\textcolor{black}{As for feature representations, the ResNet100 features
from \cite{palatucci2009zero} are trained from ImageNet 2012 dataset, which
potentially have some overlapped classes with AwA dataset. }}\textcolor{black}{. We show that, our vocabulary-informed learning
\leon{is able to} improve the recognition accuracy on all settings.}

By only using 200 training instances, we report the results on standard
supervised (on source classes), zero-shot (on target classes), and
generalized zero-shot recognition (both on source and target classes)
as shown in \leon{Table}~\ref{tab:Standard-split-on}. Note that
for ZSL and G-ZSL, our settings is a more realistic and yet more challenging
than those in previous methods \cite{romera2015embarrassingly,kodirov2017semantic},
since the source classes have few training instances. We also
compare using 100/1000-dimensional word2vec representation (\ie,
$d=100/1000$). Both the Top-1 and Top-5 classification accuracy is
reported. \textcolor{black}{Note that the key novelty of our WMM-Voc comes from
directly estimating the density of source training classes. Such
an approach would be helpful in alleviating the hubness problem and should lead to better performance in zero-shot learning. As shown in Table \ref{tab:Standard-split-on},
the improvement from MM-Voc to WMM-Voc and then further to Deep WMM-Voc
validate this point. }

We highlight the following observations: (1) Deep WMM-Voc achieves
the best zero-shot learning accuracy compared with the other state-of-art
methods. It is 18.45\% and 21.02\% higher than SAE and ESZSL respectively
on Top-1 accuracy. Our WMM-Voc can still beat the state-of-the-art
SAE and ESZSL by outperforming 17.67\% and 20.24\% individually on
Top-1 accuracy. (2) In supervised learning task, the ESZSL and Deep
WMM-Voc have almost the same performance, if we consider the variances
in sampling the 200 training instances. Our WMM-Voc is slightly better
than these two methods. (3) In G-ZSL \leon{setting}, our two models
get \leon{significantly better} 
performance compared with the other competitors. \leon{Notably, the}
Top-1 accuracy of SAE and ESZSL \leon{is} 0. While Deep WMM-Voc
and WMM-Voc both \leon{have} higher accuracy. This shows the effectiveness
of our two models. (4) As expected, the deep models that fine-tune
features along with classifiers (Deep-SVR and Deep WMM-Voc) are better
than counterparts with pre-extracted representations (SVR-Map, WMM-Voc). 

\subsubsection{Results on different training/testing splits}

We conduct experiments using different number of training
instances and compare results on tasks of supervised, zero-shot
and generalized zero-shot learning. On each split, we use both 100
and 1000 dimensional word vectors.  \textcolor{black}{We use 12,156 testing instances from source classes in 
supervised, G-ZSL and open-set setting as well as 6,180 testing instances from target classes in zero-shot, G-ZSL and open-set setting.} 
All the competitors are using the same types of features -- ResNet101. 

\begin{figure}
\begin{centering}
\includegraphics[scale=0.35]{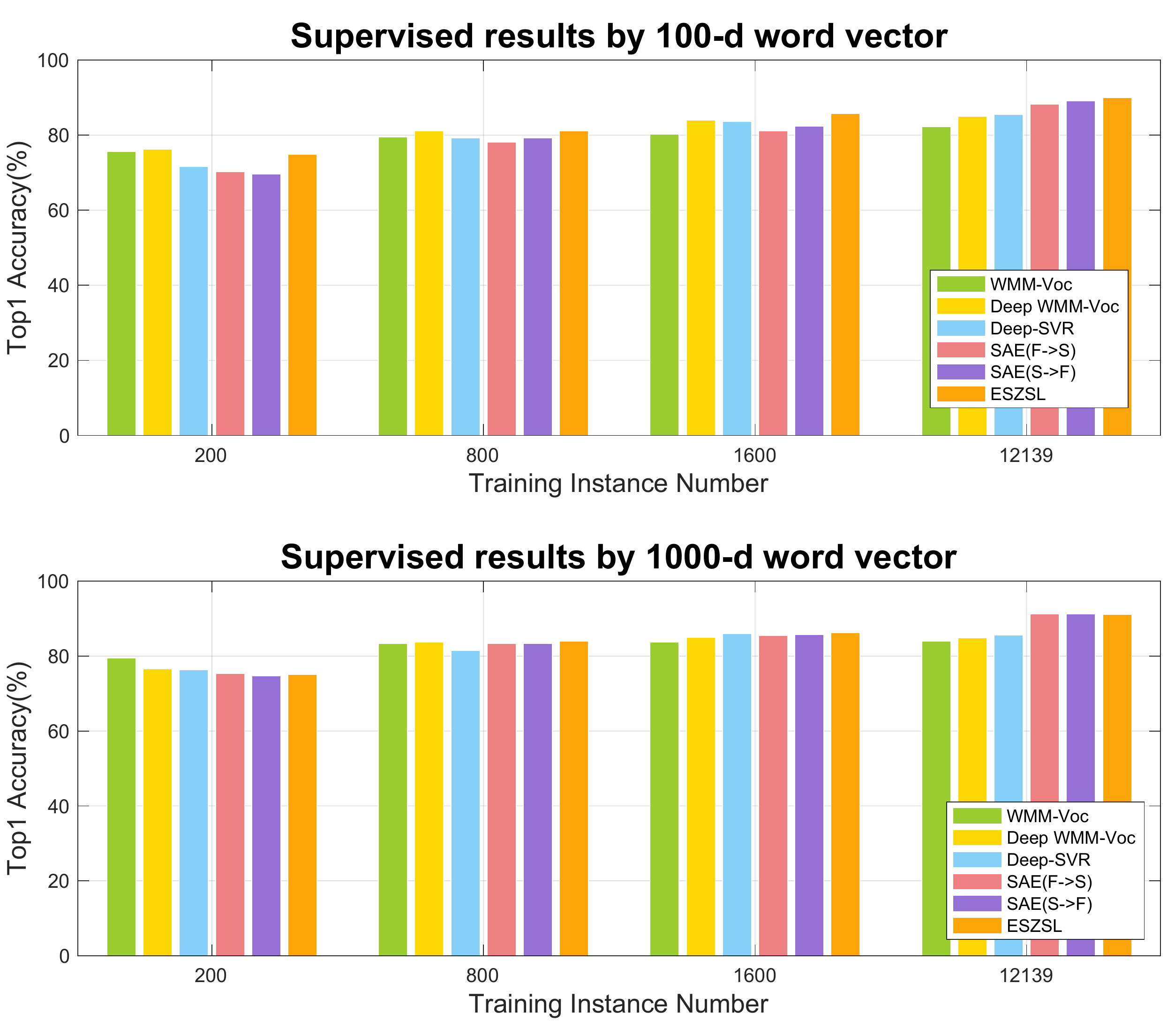} 
\par\end{centering}
\caption{\label{fig:AwA-spv-results.}Supervised learning results of AwA datasets.}
\end{figure}

\noindent \textbf{Supervised learning:} The results are compared in
\leon{Figure} \ref{fig:AwA-spv-results.}. As shown in the figure,
we observe that  \textit{\textcolor{black}{\emph{our method shows
significant improvements over the competitors in few-shot setting;
however, as the number of instance increasing, the visual semantic
mapping, $g(\mathbf{x})$, can be well learned, and the effects of additional
vocabulary-informed constraints, become less pronounced.}}}\emph{
}


\noindent \textbf{Zero-shot learning:} The ZSL results are compared
in Figure \ref{fig:AwA-ZSL-results.}. On all the settings, our two
Voc methods -- Deep WMM-Voc and WMM-Voc outperforms all the other
baselines. This validates the importance of information learning from
the open vocabulary. Further, we compare our results with the state-of-the-art
ZSL results on AwA dataset in \leon{Table}~\ref{tab:Zero-shot-Learning-Comparison}.
Our two models achieve $90.79\%$ and $90.65\%$ accuracy, which \leon{is
markedly} higher than all previous methods. This is particularly
impressive, if we take into account the fact that we use only a semantic
space and no additional attribute representations (unlike many other
competitor methods). 
We argue that much of our success and improvement
comes from a more discriminative information obtained using the open
set vocabulary and corresponding large margin constraints, rather
than from the features.
\textcolor{black}{Varying the number of training instances may slightly affect accuracy of methods reported in Table~\ref{tab:Zero-shot-Learning-Comparison}. Therefore we report the best results of each competitor and our own method at different number of training instances $\{200, 800, 1600, 121319, 24295\}$.}All competing methods in \leon{Figure} \ref{fig:AwA-ZSL-results.} \leon{use} the same
features.


\noindent \textbf{\textcolor{black}{General zero-shot learning:}}\textcolor{black}{.
The general zero-shot learning results are compared in Table \ref{tab:AwA-gzsl-results}.
We consider the accuracies of both }\textcolor{black}{\small{}$\mathbb{U}\rightarrow\mathbb{T}$
(}\textsc{\textcolor{black}{Zero-shot}}\textcolor{black}{-like}\textcolor{black}{\small{})
and $\mathbb{S}\rightarrow\mathbb{T}$ (}\textsc{\textcolor{black}{Supervised}}\textcolor{black}{-like).
In term of the harmonic mean }\textcolor{black}{\emph{\small{}$H$}}\textcolor{black}{,
our methods have significantly better performance in the general zero-shot
setting. This again shows that our framework can have better generalization
by learning from the open vocabulary.} \textcolor{black}{On the other hand, in the terms of  Area Under Seen-Unseen accuracy Curve (\emph{\small{}$AUSUC$}), the performance of Deep-SVR is very weak and the scores of ESZSL and SAE are lower than our method.  Overall, the results of \emph{\small{}$AUSUC$} still support the superiority of our methods on G-ZSL tasks. Notably, since the source domain only have 24295 instances (including training and testing images), we are unable to obtain the results of \textsc{Supervised}-like setting ($\mathbb{S}\rightarrow\mathbb{T}$), \emph{\small{}$H$} and \emph{\small{}$AUSUC$} with all source instances.}

\begin{table*}
\caption{\label{tab:AwA-gzsl-results}The G-ZSL results (100-dim/1000-dim)
of AwA dataset. We compare the results by varying the number of training instances (\emph{No.of Tr. Ins.}) of each class.  \textcolor{black}{Where $H$ is defined in Eq.~(\ref{eq:harmonic}) without calibrated stacking, $AUSUC$ means the Area Under Seen-Unseen accuracy Curve with calibrated stacking and '-' represents the unavailable results.}}
\begin{singlespace}
\noindent \begin{centering}
{\small{}}%
\begin{tabular}{c|c||cccccc}
\hline 
\multicolumn{1}{c|}{} & \textcolor{black}{\small{}Metrics}{\small{} } & \textcolor{black}{\small{}ESZSL}{\small{} } & \textcolor{black}{\small{}SAE}{\small{}} & \textcolor{black}{\small{}Deep-SVR}{\small{} } & \textcolor{black}{\small{}WMM-Voc}{\small{} } & \textcolor{black}{\small{}Deep WMM-Voc}\tabularnewline
\hline 
\hline 
\textcolor{black}{\small{}200} & \textcolor{black}{\small{}$\mathbb{U}\rightarrow\mathbb{T}$}{\small{} } & \textcolor{black}{\small{}2.88/0}{\small{} } & \textcolor{black}{\small{}2.15/0}{\small{} } & \textcolor{black}{\small{}5.65/0}{\small{} } & \textcolor{black}{\small{}28.92/27.98}{\small{} } & \textcolor{black}{\small{}33.04/}\textbf{\textcolor{black}{\small{}34.77}}\tabularnewline
 & \textcolor{black}{\small{}$\mathbb{S}\rightarrow\mathbb{T}$}{\small{} } & \textcolor{black}{\small{}75.76/76.08}{\small{} } & \textcolor{black}{\small{}70.13/75.32}{\small{} } & \textcolor{black}{\small{}71.22/}\textbf{\textcolor{black}{\small{}76.32}}{\small{} } & \textcolor{black}{\small{}70.20/74.20}{\small{} } & \textcolor{black}{\small{}71.16/69.48}\tabularnewline
 & \textcolor{black}{\emph{\small{}$H$}}{\small{} } & \textcolor{black}{\small{}5.55/0}{\small{} } & \textcolor{black}{\small{}4.17/0}{\small{} } & \textcolor{black}{\small{}10.47/0}{\small{} } & \textcolor{black}{\small{}40.96/40.64}{\small{} } & \textcolor{black}{\small{}45.13/}\textbf{\textcolor{black}{\small{}46.35}}\tabularnewline
 & \textcolor{black}{\small{}$AUSUC$}{\small{} } & \textcolor{black}{\small{}0.4231/0.4344}{\small{} } & \textcolor{black}{\small{}0.3885/0.4556}{\small{} } & \textcolor{black}{\small{}0.3048/0.3939}{\small{} } & \textcolor{black}{\small{}0.4840/\textbf{0.5190}}{\small{} } & \textcolor{black}{\small{}0.5028/0.4776}\tabularnewline
\hline 
\textcolor{black}{\small{}800}{\small{} } & \textcolor{black}{\small{}$\mathbb{U}\rightarrow\mathbb{T}$}{\small{} } & \textcolor{black}{\small{}0.19/0}{\small{} } & \textcolor{black}{\small{}0.78/0}{\small{} } & \textcolor{black}{\small{}5.34/0.02}{\small{} } & \textcolor{black}{\small{}25.57/25.68}{\small{} } & \textcolor{black}{\small{}27.59/}\textbf{\textcolor{black}{\small{}27.77}}\tabularnewline
 & \textcolor{black}{\small{}$\mathbb{S}\rightarrow\mathbb{T}$}{\small{} } & \textcolor{black}{\small{}81.14/}\textbf{\textcolor{black}{\small{}83.95}}{\small{} } & \textcolor{black}{\small{}78.02/83.41}{\small{} } & \textcolor{black}{\small{}78.92/81.46}{\small{} } & \textcolor{black}{\small{}74.23/77.33}{\small{} } & \textcolor{black}{\small{}75.53/77.19}\tabularnewline
 & \textcolor{black}{\emph{\small{}$H$}}{\small{} } & \textcolor{black}{\small{}0.38/0}{\small{} } & \textcolor{black}{\small{}1.54/0}{\small{} } & \textcolor{black}{\small{}10.00/0.04}{\small{} } & \textcolor{black}{\small{}38.04/38.56}{\small{} } & \textcolor{black}{\small{}40.42/}\textbf{\textcolor{black}{\small{}40.85}}\tabularnewline
 & \textcolor{black}{\small{}$AUSUC$}{\small{} } & \textcolor{black}{\small{}0.4409/0.4710}{\small{} } & \textcolor{black}{\small{}0.3870/0.4483}{\small{} } & \textcolor{black}{\small{}0.3452/0.4400}{\small{} } & \textcolor{black}{\small{}0.4764/\textbf{0.5387}}{\small{} } & \textcolor{black}{\small{}0.4953/0.5353}\tabularnewline
\hline 
\textcolor{black}{\small{}1600} & \textcolor{black}{\small{}$\mathbb{U}\rightarrow\mathbb{T}$}{\small{} } & \textcolor{black}{\small{}0.71/0}{\small{} } & \textcolor{black}{\small{}0.87/0}{\small{} } & \textcolor{black}{\small{}4.69/0}{\small{} } & \textcolor{black}{\small{}24.63/27.22}{\small{} } & \textbf{\textcolor{black}{\small{}33.66}}\textcolor{black}{\small{}/32.86}\tabularnewline
 & \textcolor{black}{\small{}$\mathbb{S}\rightarrow\mathbb{T}$}{\small{} } & \textcolor{black}{\small{}85.62}\textbf{\textcolor{black}{\small{}/86.24}}{\small{} } & \textcolor{black}{\small{}81.08/85.48}{\small{} } & \textcolor{black}{\small{}83.30/86.02}{\small{} } & \textcolor{black}{\small{}74.99/77.67}{\small{} } & \textcolor{black}{\small{}78.96/78.64}\tabularnewline
 & \textcolor{black}{\small{}$H$}{\small{} } & \textcolor{black}{\small{}1.41/0}{\small{} } & \textcolor{black}{\small{}1.72/0}{\small{} } & \textcolor{black}{\small{}8.88/0}{\small{} } & \textcolor{black}{\small{}37.08/40.31}{\small{} } & \textbf{\textcolor{black}{\small{}47.20}}\textcolor{black}{\small{}/46.35}\tabularnewline
 & \textcolor{black}{\small{}$AUSUC$}{\small{} } & \textcolor{black}{\small{}0.4507/0.5139}{\small{} } & \textcolor{black}{\small{}0.4190/0.4740}{\small{} } & \textcolor{black}{\small{}0.3776/0.4780}{\small{} } & \textcolor{black}{\small{}0.5016/0.5572}{\small{} } & \textcolor{black}{\small{}0.5554/\textbf{0.5733}}\tabularnewline
\hline 
\textcolor{black}{\small{}12139} & \textcolor{black}{\small{}$\mathbb{U}\rightarrow\mathbb{T}$}{\small{} } & \textcolor{black}{\small{}0.37/0}{\small{} } & \textcolor{black}{\small{}0.44/0}{\small{} } & \textcolor{black}{\small{}5.19/0}{\small{} } & \textcolor{black}{\small{}27.80/30.53}{\small{} } & \textbf{\textcolor{black}{\small{}32.23}}\textcolor{black}{\small{}/28.19}\tabularnewline
 & \textcolor{black}{\small{}$\mathbb{S}\rightarrow\mathbb{T}$}{\small{} } & \textcolor{black}{\small{}89.98/91.16}{\small{} } & \textcolor{black}{\small{}88.18/}\textbf{\textcolor{black}{\small{}91.26}}{\small{} } & \textcolor{black}{\small{}85.37/85.64}{\small{} } & \textcolor{black}{\small{}77.36/78.34}{\small{} } & \textcolor{black}{\small{}80.64/78.32}\tabularnewline
 & \textcolor{black}{\small{}$H$}{\small{} } & \textcolor{black}{\small{}0.74/0}{\small{} } & \textcolor{black}{\small{}0.88/0}{\small{} } & \textcolor{black}{\small{}9.79/0}{\small{} } & \textcolor{black}{\small{}40.90/43.94}{\small{} } & \textbf{\textcolor{black}{\small{}46.05}}\textcolor{black}{\small{}/41.46}\tabularnewline
 & \textcolor{black}{\small{}$AUSUC$}{\small{} } & \textcolor{black}{\small{}0.5096/0.5294}{\small{} } & \textcolor{black}{\small{}0.4493/0.5120}{\small{} } & \textcolor{black}{\small{}0.3353/0.4397}{\small{} } & \textcolor{black}{\small{}0.5144/0.5319}{\small{} } & \textcolor{black}{\small{}\textbf{0.5525}/0.5394}\tabularnewline
\hline 
\textcolor{black}{\small{}24295} & \textcolor{black}{\small{}$\mathbb{U}\rightarrow\mathbb{T}$}{\small{} } & \textcolor{black}{\small{}0.83/0}{\small{} } & \textcolor{black}{\small{}0.37/0}{\small{} } & \textcolor{black}{\small{}5.39/0}{\small{} } & \textcolor{black}{\small{}27.15/29.42}{\small{} } & \textbf{\textcolor{black}{\small{}35.65}}\textcolor{black}{\small{}/31.78}\tabularnewline
 & \textcolor{black}{\small{}$\mathbb{S}\rightarrow\mathbb{T}$}{\small{} } & \textcolor{black}{\small{}-}{\small{} } & \textcolor{black}{\small{}-}{\small{} } & \textcolor{black}{\small{}-} & \textcolor{black}{\small{}-}{\small{} } & \textcolor{black}{\small{}-}\tabularnewline
 & \textcolor{black}{\small{}$H$}{\small{} } & \textcolor{black}{\small{}-}{\small{} } & \textcolor{black}{\small{}-}{\small{} } & \textcolor{black}{\small{}-}{\small{} } & \textcolor{black}{\small{}-}{\small{} } & \textcolor{black}{\small{}-}\tabularnewline
 & \textcolor{black}{\small{}$AUSUC$}{\small{} } & \textcolor{black}{\small{}-}{\small{} } & \textcolor{black}{\small{}-}{\small{} } & \textcolor{black}{\small{}-}{\small{} } & \textcolor{black}{\small{}-}{\small{} } & \textcolor{black}{\small{}-}\tabularnewline
\hline 
\end{tabular}{\small\par}
\par\end{centering}
\end{singlespace}
\end{table*}

\subsubsection{Large-scale open set recognition}

\begin{figure}
\begin{centering}
\begin{tabular}{c}
\begin{tabular}{c|cccc}
\hline 
 & \multicolumn{3}{c}{Testing Classes} & \tabularnewline
\hline 
\hline 
AwA dataset  & Aux.  & Targ.  & Total  & Vocab\tabularnewline
\hline 
\hline 
\textsc{Open-Set}$_{310K}$  & (left)  & (right)  & 40 / 10  & 310K\tabularnewline
\hline 
\end{tabular}\tabularnewline
\includegraphics[scale=0.4]{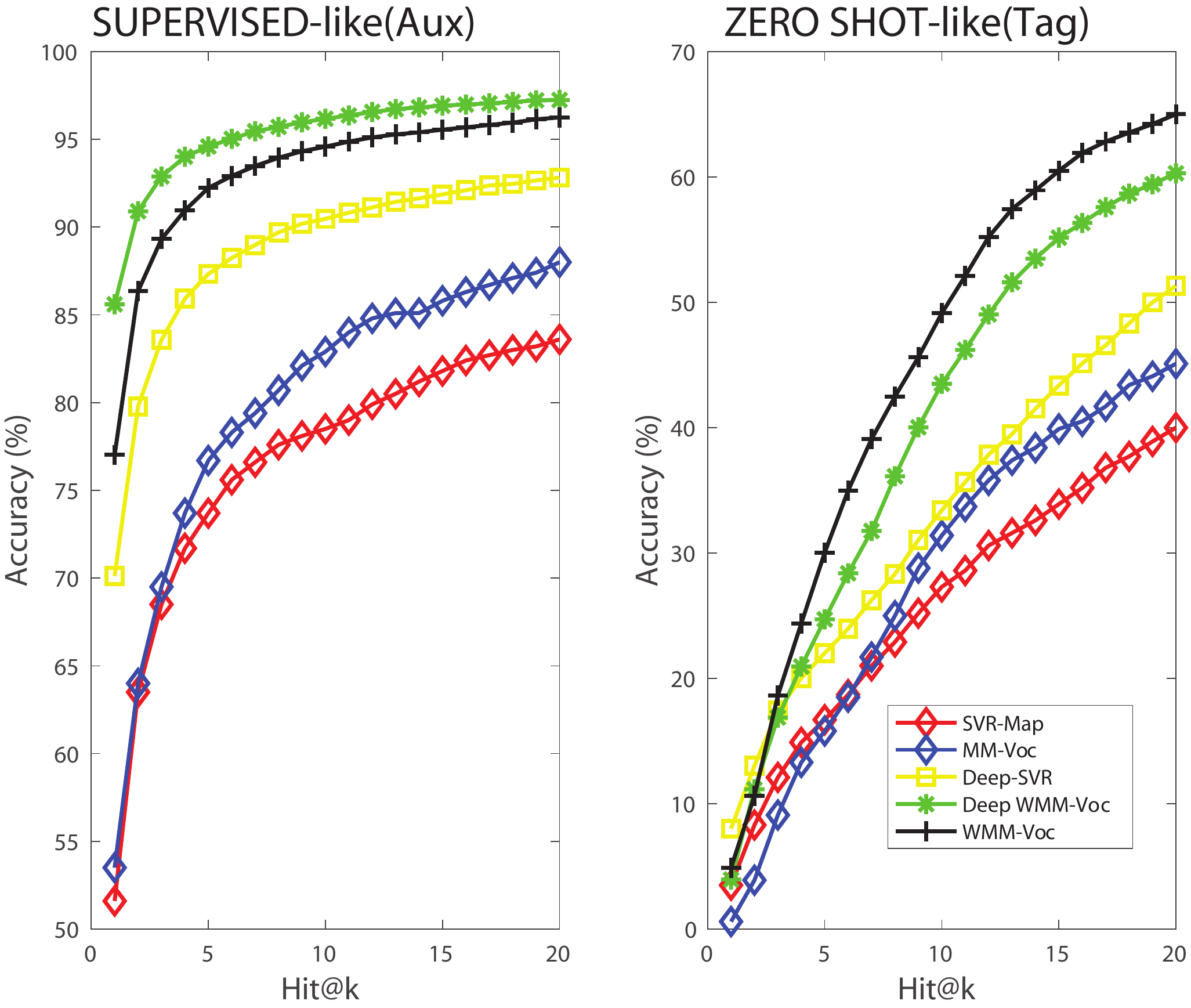}\tabularnewline
\end{tabular}
\par\end{centering}
\caption{\label{fig:awa-openset}Openset results of AwA datasets. \textcolor{black}{We use 1600 training instances equally sampled from all source classes to train the model. }}
\end{figure}

We also compare the results on \textsc{Open-set}$_{310K}$ setting
with the large vocabulary of approximately 310K entities; as such
the chance performance is much lower. We use 100-dim word
vector representations as the semantic space. While our \textsc{Open-set}
variants do not assume that test data comes from either source/auxiliary
domain or target domain, we split the two cases to mimic \textsc{Supervised}
and \textsc{Zero-shot} scenarios for easier analysis. The results
are shown in Figure \ref{fig:awa-openset}.

On \textsc{Supervised}-like setting, Figure~\textcolor{black}{\ref{fig:awa-openset}}
(left), our Deep WMM-Voc and WMM-Voc have better performance than
the other baselines. The better results are largely due to the better
embedding matrix $W$ learned by enforcing maximum margins between
training class name and open set vocabulary on source training data.
This validates the effectiveness of proposed framework. In particular,
we find that (1) The ``deep'' version always has better performance
than their corresponding ``non-deep'' \leon{counterparts}. For
example, the Deep-SVR and Deep WMM-Voc achieve higher open-set recognition
accuracy than SVR-Map and WMM-Voc. (2) The WMM-Voc has better performance
than \leon{MM-Voc}; this shows that the weighting strategy introduced
in Section \ref{subsec:Extreme-Value-Voc} can indeed help better learn
the embedding from visual to semantic space.

On \textsc{Zero shot}-like setting, our method still has a notable
advantage over that of SVR-Map, Deep-SVR methods on Top-$k$ ($k>3$)
accuracy, again thanks to the better embedding $W$ learned by Eq.~(\ref{eq:formulation}).
However, we notice that our top-1 accuracy on \textsc{Zero shot}-like
setting is lower than Deep SVR method. We find that our method tends
to label some instances from target data with their nearest classes
from within source label set. For example, ``humpback whale'' from
testing data is more likely to be labeled as ``blue whale''. However,
when considering Top-$k$ ($k>3$) accuracy, our method still has
advantages over baselines. \textcolor{black}{It suggests that the semantic embeddings may be suffering from the problem that density of source classes is more concentrated than that of target classes. To show the effectiveness of WMM-Voc, as opposed to MM-Voc, we employ the False Positive Rate as the metric, $r_{fp}=N_{e}/{N_{un}}$, where $N_{e}$
means the number of testing unseen instances predicted as seen ones
and $N_{un}$ defines the number of testing unseeen instances. Experiments are conducted on AwA dataset with all training instances, and 100-dim word vector prototypes. 
The false positive rates are $0.16$, $0.10$, $0.12$, $0.05$ and $0.06$ by using SVR, Deep-SVR, MM-Voc, WMM-Voc and Deep WMM-Voc, respectively. They further validates that WMM-Voc outperforms MM-Voc. 
}

\begin{table*}[!th]
{\tiny{}\caption{{\small{}\label{tab:Imagenet-2012/2010-dataset 3 instances.}The classification
accuracy (Top-1 / Top-5) of ImageNet $2012$/$2010$ dataset on ZERO-SHOT
and SUPERVISED settings using 3000 source training instances.}}
}{\tiny\par}
\centering{}{\small{}}%
\begin{tabular}{c|c|c|c|c||c|c|c}
\hline 
{\small{}Settings } & {\small{}SVR\leon{-Map} } & {\small{}Deep-SVR } & {\small{}ESZSL } & {\small{}SAE } & {\small{}{}\leon{}\textbf{\small{}MM-Voc}{\small{}} } & \textbf{\small{}WMM-Voc}{\small{} } & \textbf{\small{}Deep WMM-Voc}{\small{} }\tabularnewline
\hline 
\hline 
{\small{}Supervised } & {\small{}25.6/-- } & {\small{}31.26/50.51 } & {\small{}38.26/64.38 } & {\small{}32.95/54.44 } & {\small{}37.1/62.35 } & {\small{}35.95/62.77 } & \textbf{\small{}38.92/65.35}{\small{} }\tabularnewline
\hline 
{\small{}Zero-shot } & {\small{}4.1/-- } & {\small{}5.29/13.32 } & {\small{}5.86/13.71 } & {\small{}5.11/12.62 } & {\small{}8.90/14.90 } & {\small{}8.50/20.73 } & \textbf{\small{}9.26/21.99}{\small{} }\tabularnewline
\hline 
\end{tabular}{\small\par}
\end{table*}

\subsection{Experimental results on ImageNet dataset}

We validate our methods on large-scale ImageNet 2012/2010 dataset;
the 1000-dimensional word2vec representation is used here since this
dataset has larger number of classes than AwA. \textcolor{black}{The instances of testing classes are equally sampled; making experiment less sensitive to the problem of unbalanced data. To be specific, $50\times1,000$ testing instances from source classes are used in supervised, G-ZSL and open-set setting as well as $360\times100$ testing instances from target classes are used in zero-shot, G-ZSL and open-set setting.} The VGG-19 features
of ImageNet \leon{pre-trained network} 
are utilized as the input of all algorithms to make a fair comparison.
We employ the Deep-SVR, SAE, ESZSL as baselines under the \textsc{Supervised},
\textsc{Zero-shot} and \textsc{General Zero-shot} settings respectively.

\subsubsection{\emph{Pseudo}-few-shot Source Training instances }
\textcolor{black}{
The standard few-shot learning assumes disjoint instance set  on source and target domains, as discussed in Sec.~\ref{sec:few-shot-target}.  As an ablation study, we would like to simulate a few-shot-like learning task on source domain by slightly violating the standard few-shot learning assumption. We name this setting ``\emph{Pseudo}-few-shot learning'': only few source training instances are used here and the feature extractor -- VGG-19 model is pre-trained on ILSVRC 2012 dataset \cite{chatfield2014return}. The ``Pseudo-'' here indicates that large amount of instances are used to train the feature extractor, but not used in training classifiers. }
Thus the experiments in this section can be served as an additional ablation study to reveal the insights of our model in addressing the few-shot-like task on source domain.  Particularly, we conduct the experiments of using few-shot source training instance, \emph{i.e.}, 3,000 training instances used here. \leon{The} results are listed in \leon{Table}~\ref{tab:Imagenet-2012/2010-dataset 3 instances.}. We introduce this setting
to particularly focus on learning from few training samples per class,
in order to mimic human capability and performance in learning from
few examples. \leon{We} show that our vocabulary-informed learning
framework \leon{enables} learning with little data. 
\textcolor{black}{In particular, we highlight that the Top-5 performance
of WMM-Voc is much higher (>5\%) than that of MM-Voc, despite
the slightly worse performance on Top-1 accuracy. Note that the degradation
of Top-1 results on ImageNet is also understandable. Note, WMM-Voc
is only fitting the 3000 training instances on ImageNet dataset, and
the features of these training instances may not be fine-tuned/optimized
for the newly introduced penalty term of WMM-Voc. Once the features
of training instances are fine-tuned by the deep version; we can show
that the Deep WMM-Voc can improve from MM-Voc and WMM-Voc. } 

Critically, with different settings in \leon{Table} \ref{tab:Imagenet-2012/2010-dataset 3 instances.},
our vocabulary-informed learning can beat the other baselines under
\leon{all} settings. We highlight several \leon{findings}:

(1) The supervised performance of our \leon{methods} stands out
\leon{from} the state-of-art. Specifically, our Deep WMM-Voc \leon{achieves
the highest supervised recognition accuracy, with ESZSL following
closely}. \leon{SVR-Map appears to be the worst.} 

(2) On Zero-shot learning task, our proposed Deep WMM-Voc \leon{gets}
9.26\% Top-1 and 21.99\% Top-5 accuracy. It outperforms all the other
baselines. Comparing with our previous \leon{MM-Voc} result in \cite{fu2016semi},
our result is 0.36\% higher than \leon{MM-Voc}. This improvement
is statistically significant due to the few number of training instances
and large number of testing instances. Additionally, the Top-1 result
of WMM-Voc is 8.5\% which is also comparable to that of \leon{MM-Voc},
and far higher than those of SVR, Deep-SVR, ESZSL and SAE. This validates
the effectiveness of learning from \leon{open} vocabulary proposed
in our two \leon{variants}. 

(3) In G-ZSL setting, we observe that both Deep WMM-Voc and WMM-Voc
outperform all the other baselines. The full set of experiments on
G-ZSL under different settings are \leon{reported} in Table \ref{tab:ImageNet-gzsl-results}.

\begin{figure}
\begin{centering}
\includegraphics[scale=0.4]{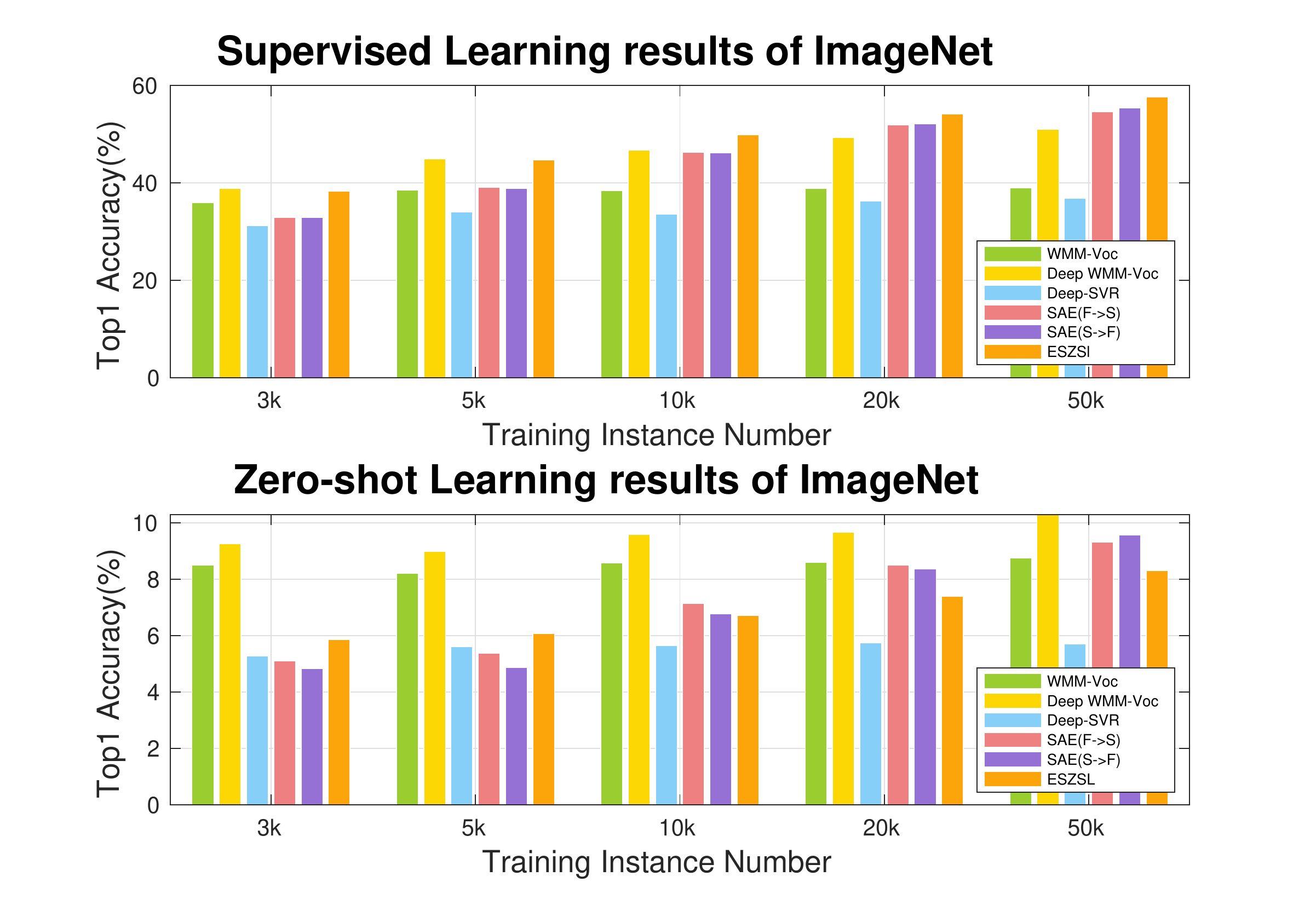} 
\par\end{centering}
\caption{\label{fig:imagenet-zsl-results.}The supervised and zero-shot learning results on ImageNet 2012/2010 dataset. }
\end{figure}

\begin{table}
\caption{\label{tab:ImageNet-gzsl-results} The G-ZSL results (1000-dim) of
ImageNet datasets. We compare the results of using different number
of training instances (No.) D-SVR, W-V and D-W-V indicate
Deep SVR, WWM-Voc, and Deep WWM-Voc, respectively.}
\begin{singlespace}
\noindent \begin{centering}
\textcolor{black}{\small{}}%
\begin{tabular}{c|c|c|c|c||c|c}
\hline 
\multicolumn{1}{c|}{\textcolor{black}{\small{}No.}} & Metrics & \textcolor{black}{\small{}ESZSL} & \textcolor{black}{\small{}SAE} & \textcolor{black}{\small{}D-SVR} & \textcolor{black}{\small{}W-V} & \textcolor{black}{\small{}D-W-V}\tabularnewline
\hline 
\hline 
\multirow{3}{*}{\begin{turn}{90}\textcolor{black}{\small$3000$}\end{turn}} & \textcolor{black}{\small{}$\mathbb{U}\rightarrow\mathbb{T}$} & \textcolor{black}{\small{}0.46} & \textcolor{black}{\small{}0.24} & \textcolor{black}{\small{}0.20} & \textbf{\textcolor{black}{\small{}2.02}} & \textcolor{black}{\small{}1.93}\tabularnewline
 & \textcolor{black}{\small{}$\mathbb{S}\rightarrow\mathbb{T}$} & \textbf{\textcolor{black}{\small{}38.07}} & \textcolor{black}{\small{}32.86} & \textcolor{black}{\small{}31.06} & \textcolor{black}{\small{}32.40} & \textcolor{black}{\small{}36.61}\tabularnewline
 & \textcolor{black}{\small{}$H$} & \textcolor{black}{\small{}0.91} & \textcolor{black}{\small{}0.48} & \textcolor{black}{\small{}0.40} & \textbf{\textcolor{black}{\small{}3.80}} & \textcolor{black}{\small{}3.67}\tabularnewline
\hline 
\multirow{3}{*}{\begin{turn}{90}\textcolor{black}{\small{}$10000$}\end{turn}} & \textcolor{black}{\small{}$\mathbb{U}\rightarrow\mathbb{T}$} & \textcolor{black}{\small{}0.38} & \textcolor{black}{\small{}0.18}  & \textcolor{black}{\small{}0.18} & \textbf{\textcolor{black}{\small{}2.01}} & \textcolor{black}{\small{}1.99}\tabularnewline
 & \textcolor{black}{\small{}$\mathbb{S}\rightarrow\mathbb{T}$} & \textbf{\textcolor{black}{\small{}49.65}} & \textcolor{black}{\small{}46.23} & \textcolor{black}{\small{}33.54} & \textcolor{black}{\small{}32.87} & \textcolor{black}{\small{}43.53}\tabularnewline
 & \textcolor{black}{\small{}$H$} & \textcolor{black}{\small{}0.75} & \textcolor{black}{\small{}0.36} & \textcolor{black}{\small{}0.36} & \textcolor{black}{\small{}3.79} & \textbf{\textcolor{black}{\small{}3.81}}\tabularnewline
\hline 
\multirow{3}{*}{\begin{turn}{90}\textcolor{black}{\small{}$50000$}\end{turn}} & \textcolor{black}{\small{}$\mathbb{U}\rightarrow\mathbb{T}$} & \textcolor{black}{\small{}0.37} & \textcolor{black}{\small{}0.19} & \textcolor{black}{\small{}0.20} & \textcolor{black}{\small{}2.11} & \textbf{\textcolor{black}{\small{}2.15}}\tabularnewline
 & \textcolor{black}{\small{}$\mathbb{S}\rightarrow\mathbb{T}$} & \textbf{\textcolor{black}{\small{}57.43}} & \textcolor{black}{\small{}54.55} & \textcolor{black}{\small{}36.75} & \textcolor{black}{\small{}33.16} & \textcolor{black}{\small{}47.28}\tabularnewline
 & \textcolor{black}{\small{}$H$} & \textcolor{black}{\small{}0.74} & \textcolor{black}{\small{}0.38} & \textcolor{black}{\small{}0.40} & \textcolor{black}{\small{}3.97} & \textbf{\textcolor{black}{\small{}4.11}}\tabularnewline
\hline 
\end{tabular}{\small\par}
\par\end{centering}
\end{singlespace}
\end{table}

\subsubsection{Few-shot Target Training instances \label{sec:few-shot-target}}

\textcolor{black}{We further introduce few-shot learning experiments
on target instances to validate the performance of our methods. The
experiments are conducted on ImageNet dataset. In total, there are
360 target classes from ImageNet 2010 data split with 100 instances
per class; the feature extractor -- VGG-19 is trained on the 1000
classes from ImageNet 2012. The 1 or 3 training instances are sampled
from each target class. The other instances of the target classes
are utilized as the test set. This is the few-shot learning setting,
which is consistent with general definition \cite{Fei-Fei:2006:OLO:1115692.1115783}.
We compare to SVM, KNN, Deep SVR, and SAE. The
results are shown in Table~\ref{tab:results-of-few-shot-target-Imagenet}.
We can see that our method (WMM-Voc) can beat all the other competitors. 
Particularly, we have an obvious advantage in 1-shot target setting. Our Deep variant
(Deep WMM-Voc) has better performance both in 1- and 3-shot setting.
This shows the efficacy of proposed methods in few-shot learning task.}

\begin{table}
\caption{\textcolor{black}{\label{tab:results-of-few-shot-target-Imagenet}Results
of few-shot target training instances on ImageNet dataset.}}
\noindent \begin{centering}
\textcolor{black}{\small{}}%
\begin{tabular}{c|c|c}
\hline 
\textcolor{black}{\small{}Method} & \textcolor{black}{\small{}1-instance} & \textcolor{black}{\small{}3-instance}\tabularnewline
\hline 
\hline 
\textcolor{black}{\small{}SVM} & \textcolor{black}{\small{}2.65} & \textcolor{black}{\small{}9.81}\tabularnewline
\hline 
\textcolor{black}{\small{}KNN} & \textcolor{black}{\small{}5.23} & \textcolor{black}{\small{}13.3}\tabularnewline
\hline 
\textcolor{black}{\small{}Deep SVR} & \textcolor{black}{\small{}14.01} & \textcolor{black}{\small{}25.00}\tabularnewline
\hline 
\textcolor{black}{\small{}SAE} & \textcolor{black}{\small{}14.93} & \textcolor{black}{\small{}26.42}\tabularnewline
\hline 
\hline 
\textcolor{black}{\small{}WMM-Voc} & \textcolor{black}{\small{}17.26} & \textcolor{black}{\small{}26.59}\tabularnewline
\hline 
\textcolor{black}{\small{}Deep WMM-Voc} & \textbf{\textcolor{black}{\small{}17.95}} & \textbf{\textcolor{black}{\small{}30.44}}\tabularnewline
\hline 
\end{tabular}{\small\par}
\par\end{centering}
\end{table}

\subsubsection{Results on different training/testing splits }

We further validate our findings on ImageNet 2012/2010 dataset. In
general, our framework has advantages over the baselines since
open vocabulary helps \leon{inform} the learning process when
few training instances or limited training data is available. 
The results are compared in \leon{Figure} \ref{fig:imagenet-zsl-results.}.

\noindent \textbf{Supervised learning:} As shown in \leon{Figure}
\ref{fig:imagenet-zsl-results.}, we compare the supervised results
by increasing the training instances from 3,000 to 50,000. With 3,000
training instances, the results of Deep WMM-Voc are better than all
the other baselines with the help of learning from free vocabulary.
We further evaluate our models with larger number of training instances
(> 3 per class). We observe that for standard supervised learning
setting, the improvements achieved using vocabulary-informed learning
tend to somewhat diminish as the number of training instances substantially
grows. With large number of training instances, the mapping between
low-level image features and semantic words, $g(\mathbf{x})$, becomes
better behaved and effect of additional constraints, due to the open-vocabulary,
becomes less pronounced.

\noindent \textbf{Zero-shot Learning:} We further validate the results
on zero-shot learning setting. \leon{Figure} \ref{fig:imagenet-zsl-results.}
shows that our \leon{models} can beat \leon{all} other baselines.
\leon{Our} Deep WMM-Voc always performs the best with the source
training instances increased from 3,000 to 50,000. The WMM-Voc always
has the second best performance; especially when only few source training
instances are available, \ie, 3,000 and 5,000 training instances. Our
Deep WMM-Voc and WMM-Voc demonstrate \leon{significant improvements}
over the competitors in ZSL task. The good performance of Deep WMM-Voc
and WMM-Voc is largely due to our vocabulary-informed learning framework
\leon{which} can leverage the discriminative information from open
vocabulary and max-margin constraints, \leon{helping to} improve
performance.


\noindent \textbf{General Zero-shot Learning:} \leon{In G-ZSL, our}
methods still have \leon{the best performance} compared \leon{to
the baselines}, \leon{as} seen from Table \ref{tab:ImageNet-gzsl-results}.
The Top-1 results of WMM-Voc and Deep WMM-Voc are beyond 2\%; in contrast,
the performance of other state-of-art methods are lower than 0.5\%.

\begin{table}[!t]
\caption{\label{tab:imagenet-zsl.}\textbf{ImageNet comparison to state-of-the-art
on ZSL:} We compare the results of using $3,000$/{\em all} training
instances for all methods; T-1 (top 1) and T-5 (top 5) classification
in (\%) is reported. The VGG-19 features are used for all methods.}

\centering{}{\small{}}%
\begin{tabular}{c|c|c|c}
\hline 
{\small{}Methods } & {\small{}{}S. Sp } & {\small{}{}T-1 } & {\small{}{}T-5}\tabularnewline
\hline 
\hline 
\textbf{\small{}\leon{Deep WMM-Voc}}{\small{} } & {\small{}W } & \textbf{\small{}9.26/10.29}{\small{} } & {\small{}{}21.99/23.12}\tabularnewline
\hline 
\textbf{\small{}\leon{WMM-Voc}}{\small{} } & {\small{}W } & {\small{}{}8.5/8.76 } & {\small{}{}20.30/21.36}\tabularnewline
\hline 
\textbf{\small{}\leon{MM-Voc}}{\small{} } & {\small{}{}W } & {\small{}{}8.9/9.5 } & {\small{}{}14.9/16.8}\tabularnewline
\hline 
\hline 
{\small{}{SAE} } & {\small{}{}W } & {\small{}{}5.11/9.32 } & {\small{}{}12.26/21.04}\tabularnewline
\hline 
{\small{}{ESZSL} } & {\small{}{}W } & {\small{}{}5.86/8.3 } & {\small{}{}13.71/18.2}\tabularnewline
\hline 
{\small{}{Deep-SVR} } & {\small{}{}W } & {\small{}{}5.29/5.7 } & {\small{}{}13.32/14.12}\tabularnewline
\hline 
{\small{}Embed \cite{zhang2017learning} } & {\small{}W } & {\small{}--/}\textbf{\small{}11.00}{\small{} } & {\small{}--/}\textbf{\small{}25.70}\tabularnewline
\hline 
{\small{}{}ConSE \cite{norouzi2013zero} } & {\small{}{}W } & {\small{}{}5.5/7.8 } & {\small{}{}13.1/15.5}\tabularnewline
\hline 
{\small{}{}DeViSE \cite{frome2013devise} } & {\small{}{}W } & {\small{}{}3.7/5.2 } & {\small{}{}11.8/12.8}\tabularnewline
\hline 
{\small{}{}AMP \cite{fu2015zero} } & {\small{}{}W } & {\small{}{}3.5/6.1 } & {\small{}{}10.5/13.1}\tabularnewline
\hline 
{\small{}{}Chance } & {\small{}{}-- } & {\small{}{}$2.78e{\text{-}3}$ } & {\small{}{}--}\tabularnewline
\hline 
\end{tabular}{\small\par}
\end{table}


\noindent \textbf{\leon{Varying training set size:}} In \leon{Figure}~\ref{fig:imagenet-zsl-results.}
we also evaluate our model with the larger number of training instances
($>3$ per class) \leon{in all settings. The results are inline with
prior findings.} 

\noindent \textbf{The state-of-the-art on ZSL:} We compare our results
to several state-of-the-art large-scale zero-shot recognition models.
Our results are better than those of ConSE, DeViSE, Deep-SVR, SAE,
ESZSL and AMP on both T-1 and T-5 metrics with a {\em very} significant
margin. Poor results of DeViSE with $3,000$ training instances are
largely due to the inefficient learning of visual-semantic embedding
matrix. AMP algorithm also relies on the embedding matrix from DeViSE,
which explains similar poor performance of AMP with $3,000$ training
instances. \leon{Table}~\ref{tab:imagenet-zsl.} \leon{shows that}
our Deep WMM-Voc obtains \leon{good} performance with \leon{(all)}
50,000 training instances. Top-5 accuracy of our methods are beyond
20\%. \leon{This} again validates that our proposed methods can
have the advantages of learning from limited available training instances
by leveraging the discriminative information from open vocabulary.
\leon{Embed \cite{embedding_cvpr17} has slightly better ZSL performance
compared to our models.} However, unlike the other works \leon{that}
directly \leon{use} word vector \leon{representations} of class
names, \cite{embedding_cvpr17} require additional textual description\leon{s}
of each class to learn 
better 
\leon{class prototypes.} 

\begin{figure}
\begin{centering}
\begin{tabular}{c|ccc|c}
\hline 
 & \multicolumn{3}{c}{Testing Classes} & \tabularnewline
\hline 
\hline 
\textit{ImageNet Data}  & Aux.  & Tag.  & Total  & Vocab \tabularnewline
\hline 
\hline 
\textsc{Open-Set}$_{310K}$  & (left)  & (right)  & 1000~/~360  & $310K$ \tabularnewline
\hline 
\end{tabular}
\par\end{centering}
\begin{centering}
\par\end{centering}
\begin{centering}
\includegraphics[scale=0.5]{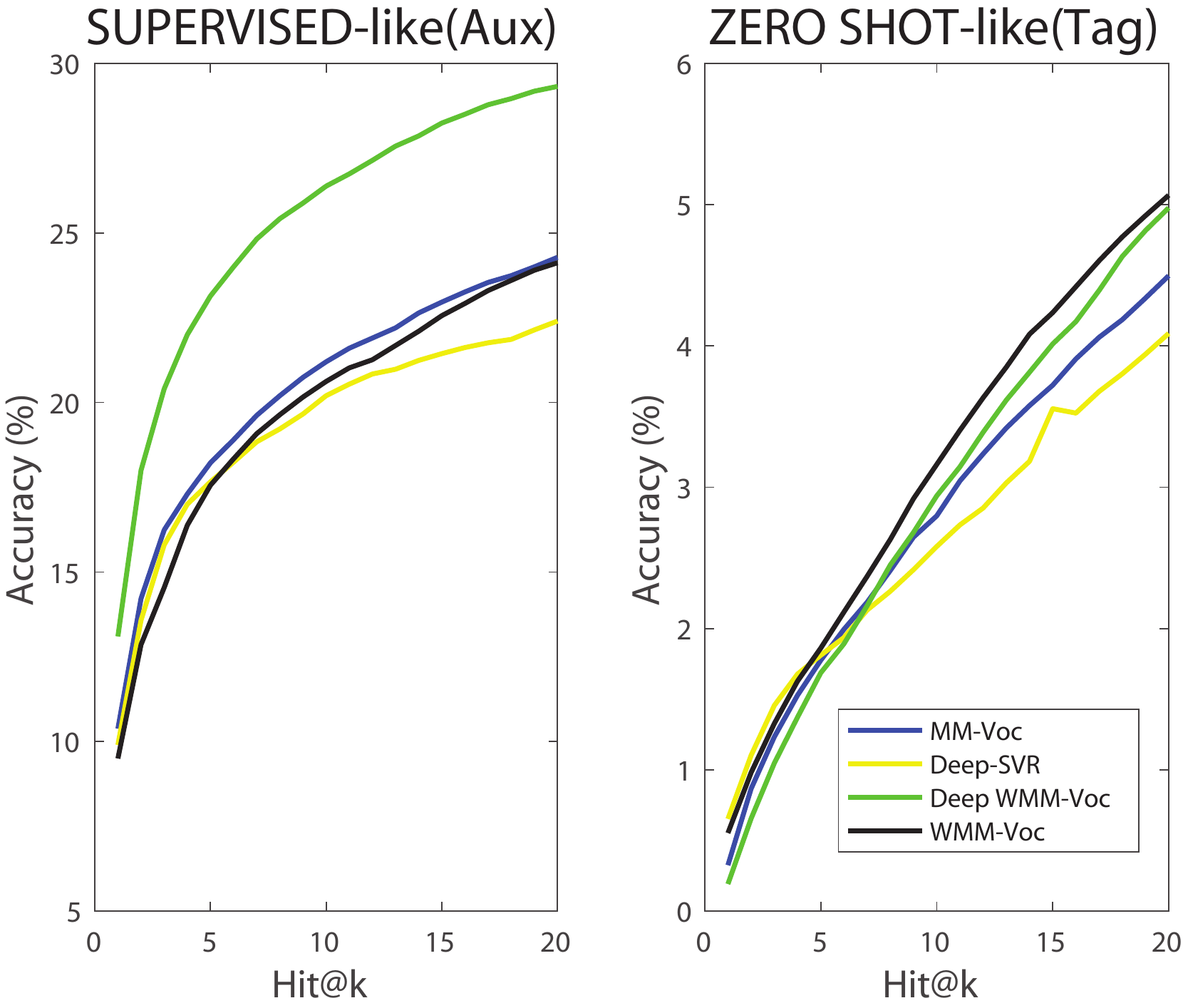} 
\par\end{centering}
\caption{\textcolor{black}{\label{fig:ImageNet-2012/2010-open} }\textbf{\textcolor{black}{Open
set recognition results on ImageNet 2012/2010 dataset:}}\textcolor{black}{{}
Openness=$0.9839$. Chance=$3.2e-4\%$. We use the synsets of each
class--- a set of synonymous (word or prhase) terms as the ground
truth names for each instance. } \textcolor{black}{We use the model trained with 50,000 instances sampled equally from source classes. }}
\end{figure}

\noindent \textbf{Open-set recognition:} The open set image recognition
results are shown in \leon{Figure}~\ref{fig:ImageNet-2012/2010-open}.
On \textsc{Supervised}-like settings, we notice the \leon{MM-Voc}
and WMM-Voc have similar open set recognition accuracy. Since this
dataset is very large, linear mapping $g(\mathbf{x})$ may not have
enough capacity to model the embedding mapping from visual space to
semantic space. Thus adding constraints on source training classes
\leon{in} WMM-Voc may slightly hinder the learning such an embedding.
That explains why the results of WMM-Voc are slightly inferior to
\leon{MM-Voc}. Deep WMM-Voc has the best performance, due \leon{to
its ability to fine-tune} low-level feature representation \leon{while
learning the embedding}. On the \textsc{Zero-shot}-like setting,
our WMM-Voc and Deep WMM-Voc have the best performance. 

\noindent 

\begin{figure}
\centering{}\includegraphics[scale=0.4]{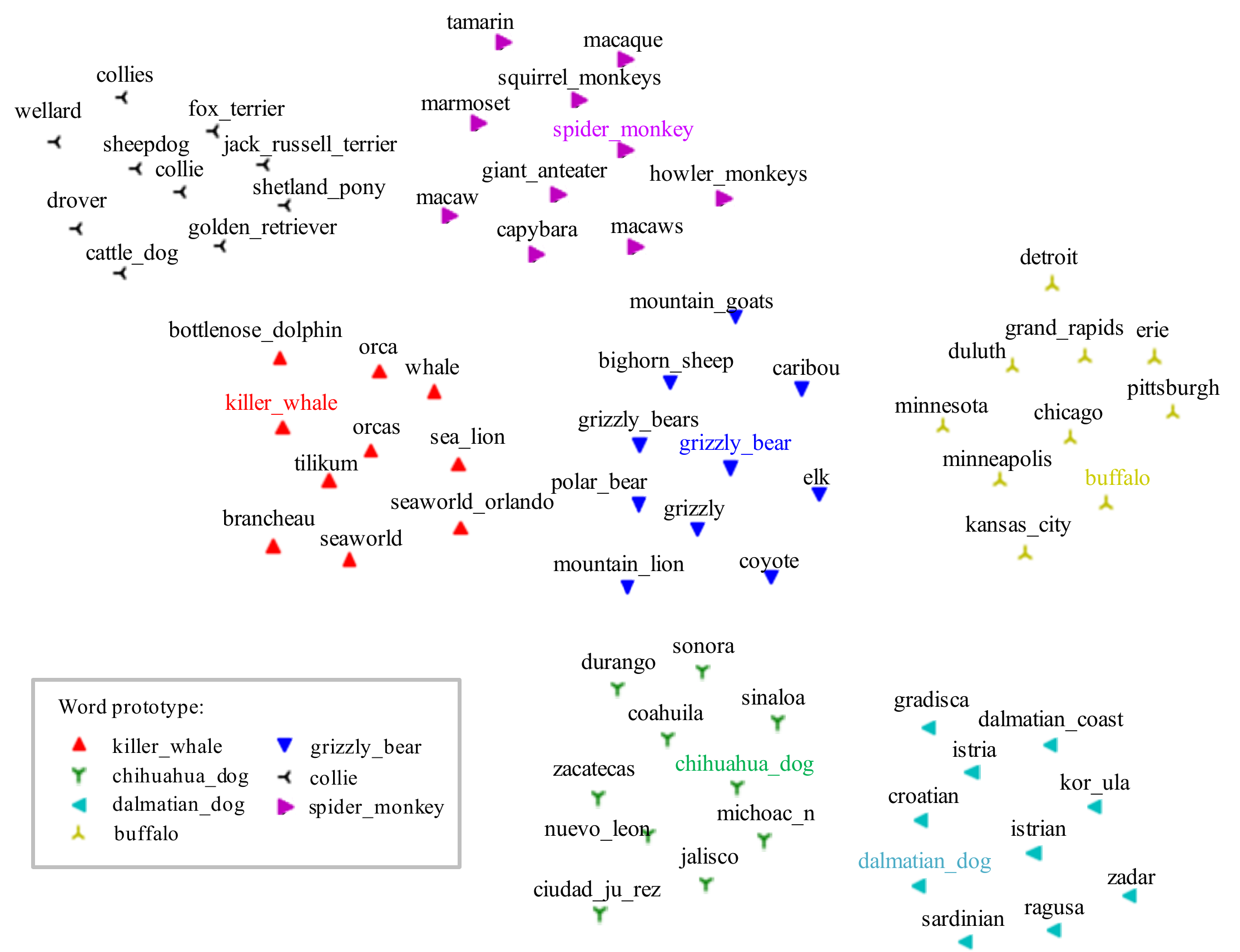}\caption{\label{fig:Visualization-of-the-semantic-space} \textbf{ Visualization of the
semantic space:} We show the t-SNE visualization of the semantic space.
The words in boxes are the mapping of training image in the semantic
space, and close neighbors are shown. The neighborhoods extend the
single training data to a space semantically meaningful.}
\end{figure}

\noindent \textbf{Qualitative visualization:} We illustrate the embedding
space learned by our Deep WMM-Voc model for the ImageNet2012/2010
dataset in Figure~\ref{fig:intro}. In particular, we have 4 source/auxiliary
and 2 target/zero-shot classes in this figure. The better separation
among classes is largely attributed to open-set max-margin constraints
introduced in our vocabulary-informed learning model. We further visualize
the semantic space in \leon{Figure} \ref{fig:Visualization-of-the-semantic-space}.
Critically, we list seven target classes on AwA dataset, as well as
their surrounding neighborhood \leon{open} vocabulary. For example,
``orcas'' is very near to ``killer\_whale''. \leon{While} 
``orcas'' \leon{are} semantically different from ``killer\_whale'',
the difference is much \leon{smaller} if we compare the ``orcas''
with the other classes, such as ``spider monkey'', ``grizzly\_bear''
and so on. \leon{Hence} the ``orcas'' can be used to help \leon{learn}
the class of ``killer\_whale'' in our vocabulary-informed learning
framework.

\section{Conclusion and Future Work}

This paper introduces the learning paradigm of vocabulary-informed
learning, by utilizing open set semantic vocabulary to help train
better classifiers for observed and unobserved classes in supervised
learning, ZSL, G-ZSL, and open set image recognition settings. We formulate
vocabulary-informed learning in the maximum margin frameworks. Extensive
experimental results illustrate the efficacy of such learning paradigm.
Strikingly, it achieves competitive performance with very few training
instances and is relatively robust to a large open set vocabulary of
up to $310,000$ class labels.

\section*{Acknowledgments}

This work was supported in part by NSFC Project (61702108, 61622204), STCSM Project (16JC1420400), Eastern Scholar (TP2017006),  Shanghai Municipal 
Science and Technology Major Project (2017SHZDZX01, 2018SHZDZX01) and ZJLab.

\bibliographystyle{abbrv}
\bibliography{egbib}


\begin{IEEEbiography}[{\includegraphics[width=1in,
height=1.25in,clip, keepaspectratio]{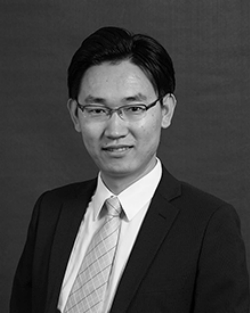}}]{Yanwei Fu} 
received the Ph.D. degree from Queen Mary University of London in 2014, and the M.Eng. degree from the Department of Computer Science and Technology, Nanjing University, China, in 2011. He held a post-doctoral position at Disney Research, Pittsburgh, PA, USA, from 2015 to 2016. He is currently a tenure-track Professor with Fudan University. His research interests are image and video understanding, and life-long learning.
\end{IEEEbiography}

\begin{IEEEbiography}[{\includegraphics[width=1in,
height=1.25in,clip, keepaspectratio]{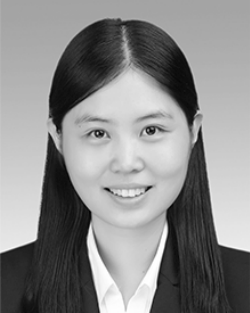}}]{Xiaomei Wang} 
is a PhD student in the School of Computer Science of Fudan University. She received the Master degree of communication and information system from Shanghai University in 2016 and the Bachelor degree of electronic information engineering from Shandong University of Technology in 2012. Her reaseach interests include zero-shot/few-shot learning, image/video captioning and visual question answering.
\end{IEEEbiography}

\begin{IEEEbiography}[{\includegraphics[width=1in,
height=1.25in,clip, keepaspectratio]{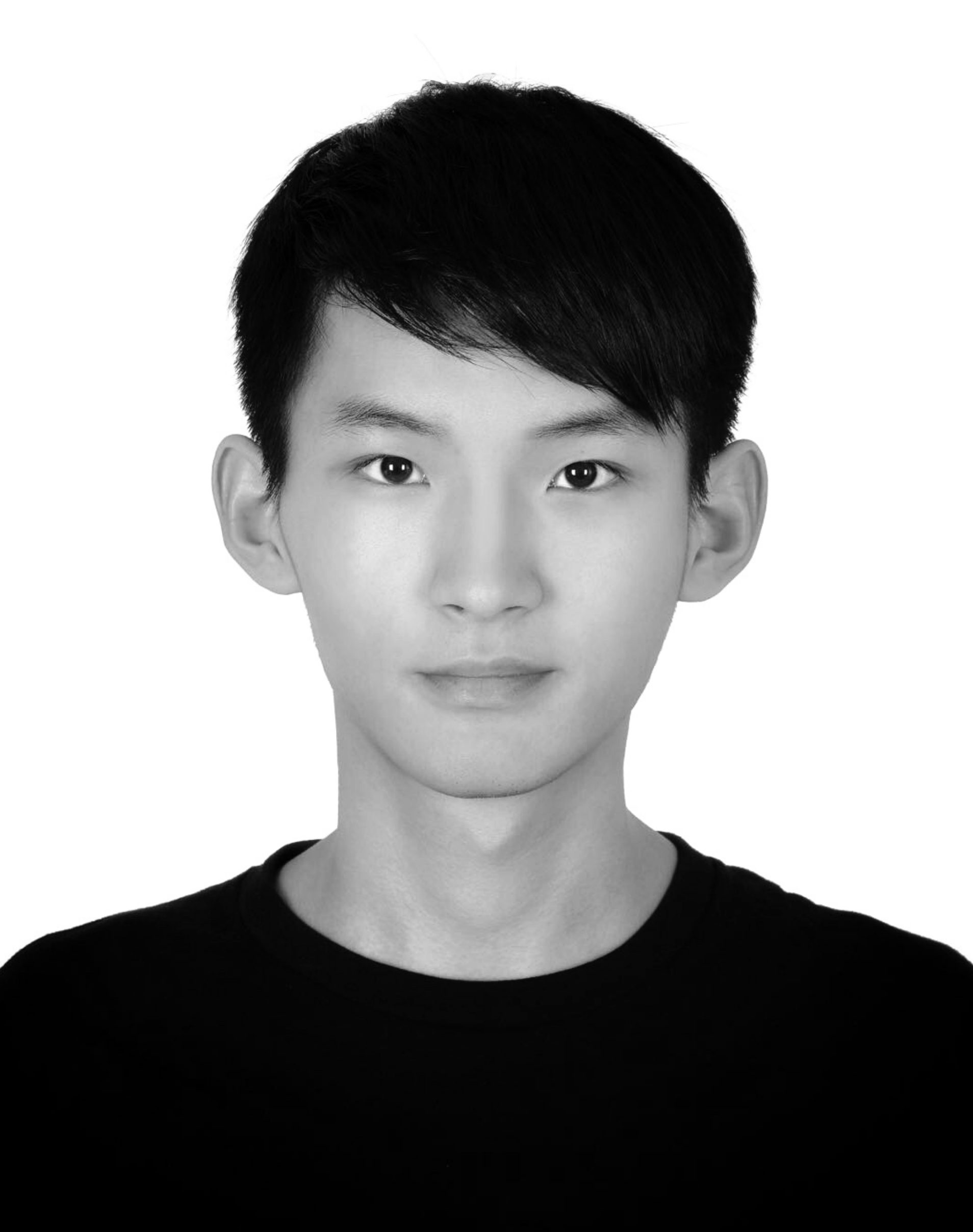}}]{Hanze Dong} 
is an undergraduate student majoring in mathematics (data science track) at the School of Data Science, Fudan University. He works in Shanghai Key Lab of Intelligent Information Processing under the supervision of Professor Yanwei Fu. His current research interests include both machine learning theory and its applications.
\end{IEEEbiography}
\begin{IEEEbiography}[{\includegraphics[width=1in,
height=1.25in,clip, keepaspectratio]{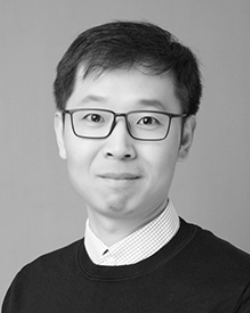}}]{Yu-Gang Jiang} is Professor of Computer Science at Fudan University and Director of Fudan-Jilian Joint Research Center on Intelligent Video Technology, Shanghai, China. He is interested in all aspects of extracting high-level information from big video data, such as video event recognition, object/scene recognition and large-scale visual search. His work has led to many awards, including the inaugural ACM China Rising Star Award, the 2015 ACM SIGMM Rising Star Award, and the research award for outstanding young researchers from NSF China. He is currently an associate editor of ACM TOMM, Machine Vision and Applications (MVA) and Neurocomputing. He holds a PhD in Computer Science from City University of Hong Kong and spent three years working at Columbia University before joining Fudan in 2011.
\end{IEEEbiography}
\enlargethispage{-2.5in}
\newpage
\begin{IEEEbiography}[{\includegraphics[width=1in,
height=1.25in,clip, keepaspectratio]{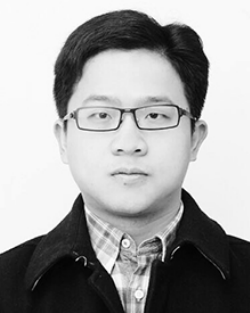}}]{Meng Wang} is a professor at the Hefei University of Technology,
China. He received his B.E. degree and Ph.D. degree in the Special
Class for the Gifted Young and the Department of Electronic
Engineering and Information Science from the University of Science and
Technology of China (USTC), Hefei, China, in 2003 and 2008,
respectively. His current research interests include multimedia
content analysis, computer vision, and pattern recognition. He has
authored more than 200 book chapters, journal and conference papers in
these areas. He is the recipient of the ACM SIGMM Rising Star Award 2014.
He is an associate editor of IEEE Transactions on Knowledge and Data
Engineering (IEEE TKDE), IEEE Transactions on Circuits and Systems
for Video Technology (IEEE TCSVT), IEEE Transactions on Multimedia (IEEE TMM), and IEEE Transactions on Neural Networks and Learning Systems (IEEE TNNLS).
\end{IEEEbiography}
\begin{IEEEbiography}[{\includegraphics[width=1in,
height=1.25in,clip, keepaspectratio]{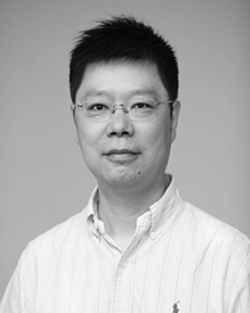}}]{Xiangyang Xue}received the BS, MS, and PhD degrees in communication engineering from Xidian University, Xi'an, China, in 1989, 1992, and 1995, respectively. He is currently a professor of computer science with Fudan University, Shanghai, China. His research interests include computer vision, multimedia information processing and machine learning.
\end{IEEEbiography}
\begin{IEEEbiography}[{{\includegraphics[clip,width=1in,height=1.25in,keepaspectratio]{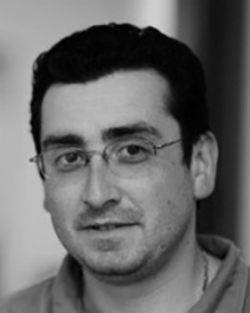}}}]{Leonid 
  Sigal} is an Associate Professor in the Department of Computer Science 
at the University of British Columbia and a Faculty Member of the Vector 
Institute for Artificial Intelligence. He is a recipient of Canada CIFAR 
AI Chair and NSERC Canada Research Chair (CRC) in Computer Vision and 
Machine Learning. Prior to this he was a Senior Research Scientist at 
Disney Research. He completed his Ph.D. at Brown University in 2008; 
received his M.A. from Boston University in 1999, and M.Sc. from Brown 
University in 2003. Leonid's research interests lie in the areas of 
computer vision, machine learning, and computer graphics. Leonid's 
research emphasis is on machine learning and statistical approaches for 
visual recognition, reasoning, understanding and analytics. He has 
published more than 70 papers in venues and journals in these fields 
(including TPAMI, IJCV, CVPR, ICCV and NeurIPS).
\end{IEEEbiography}
\enlargethispage{-3in}

\end{document}